\documentclass[draftcls,onecolumn]{IEEEtran}
\usepackage{xr}
\usepackage{cite}
\usepackage{amsmath}
\usepackage[colorlinks=true,linkcolor=black,citecolor=black,urlcolor=black]{hyperref}
\usepackage{cases}
\usepackage[utf8]{inputenc}
\usepackage[english]{babel}
\usepackage{footnote}
\usepackage{booktabs}
\usepackage{bm}
\usepackage{caption}

\usepackage[mathscr]{eucal}
\usepackage{epsfig,epsf,psfrag}
\usepackage{amssymb,amsmath,amsfonts,latexsym}
\usepackage{amsmath,graphicx,bm,xcolor,url}
\usepackage[caption=false]{subfig} 
\usepackage{fixltx2e}
\usepackage{array}
\usepackage{verbatim}
\usepackage{bm}
\usepackage{algpseudocode}
\usepackage{algorithm}
\usepackage{verbatim}
\usepackage{textcomp}
\usepackage{mathrsfs}
\usepackage{epstopdf}
\usepackage{relsize}
\usepackage{cleveref} 
\usepackage{subfig}
 \usepackage{amsthm}

\catcode`~=11 \def\UrlSpecials{\do\~{\kern -.15em\lower .7ex\hbox{~}\kern .04em}} \catcode`~=13 

\allowdisplaybreaks[3]

\DeclareMathAlphabet{\mathbsf}{OT1}{cmss}{bx}{n}
\DeclareMathAlphabet{\mathssf}{OT1}{cmss}{m}{sl}

\DeclareSymbolFont{bsfletters}{OT1}{cmss}{bx}{n}  
\DeclareSymbolFont{ssfletters}{OT1}{cmss}{m}{n}
\DeclareMathSymbol{\bsfGamma}{0}{bsfletters}{'000}
\DeclareMathSymbol{\ssfGamma}{0}{ssfletters}{'000}
\DeclareMathSymbol{\bsfDelta}{0}{bsfletters}{'001}
\DeclareMathSymbol{\ssfDelta}{0}{ssfletters}{'001}
\DeclareMathSymbol{\bsfTheta}{0}{bsfletters}{'002}
\DeclareMathSymbol{\ssfTheta}{0}{ssfletters}{'002}
\DeclareMathSymbol{\bsfLambda}{0}{bsfletters}{'003}
\DeclareMathSymbol{\ssfLambda}{0}{ssfletters}{'003}
\DeclareMathSymbol{\bsfXi}{0}{bsfletters}{'004}
\DeclareMathSymbol{\ssfXi}{0}{ssfletters}{'004}
\DeclareMathSymbol{\bsfPi}{0}{bsfletters}{'005}
\DeclareMathSymbol{\ssfPi}{0}{ssfletters}{'005}
\DeclareMathSymbol{\bsfSigma}{0}{bsfletters}{'006}
\DeclareMathSymbol{\ssfSigma}{0}{ssfletters}{'006}
\DeclareMathSymbol{\bsfUpsilon}{0}{bsfletters}{'007}
\DeclareMathSymbol{\ssfUpsilon}{0}{ssfletters}{'007}
\DeclareMathSymbol{\bsfPhi}{0}{bsfletters}{'010}
\DeclareMathSymbol{\ssfPhi}{0}{ssfletters}{'010}
\DeclareMathSymbol{\bsfPsi}{0}{bsfletters}{'011}
\DeclareMathSymbol{\ssfPsi}{0}{ssfletters}{'011}
\DeclareMathSymbol{\bsfOmega}{0}{bsfletters}{'012}
\DeclareMathSymbol{\ssfOmega}{0}{ssfletters}{'012}

\newcommand{\btheta}{\bm{\theta}}

\newcommand{\qednew}{\nobreak \ifvmode \relax \else
      \ifdim\lastskip<1.5em \hskip-\lastskip
      \hskip1.5em plus0em minus0.5em \fi \nobreak
      \vrule height0.75em width0.5em depth0.25em\fi}

\begin{document}
    
\title{Diffusion Model Based Signal Recovery \\ Under 1-Bit Quantization}

\author{Youming Chen and Zhaoqiang Liu$^{\ast}$\thanks{$^{\ast}$Corresponding author.} 
    
    \thanks{The authors are with the School of Computer Science and Engineering, University of Electronic Science and Technology of China.}}

\maketitle

\begin{abstract}
    Diffusion models (DMs) have demonstrated to be powerful priors for signal recovery, but their application to 1-bit quantization tasks, such as 1-bit compressed sensing and logistic regression, remains a challenge. This difficulty stems from the inherent non-linear link function in these tasks, which is either non-differentiable or lacks an explicit characterization. To tackle this issue, we introduce Diff-OneBit, which is a fast and effective DM-based approach for signal recovery under 1-bit quantization. Diff-OneBit addresses the challenge posed by non-differentiable or implicit links functions via leveraging a differentiable surrogate likelihood function to model 1-bit quantization, thereby enabling gradient based iterations. This function is integrated into a flexible plug-and-play framework that decouples the data-fidelity term from the diffusion prior, allowing any pretrained DM to act as a denoiser within the iterative reconstruction process. Extensive experiments on the FFHQ, CelebA and ImageNet datasets demonstrate that Diff-OneBit gives high-fidelity reconstructed images, outperforming state-of-the-art methods in both reconstruction quality and computational efficiency across 1-bit compressed sensing and logistic regression tasks. Our code is available at \url{https://github.com/Chenyouming123/DiffOneBit}.
\end{abstract}

\section{Introduction}\label{sec:introduction}

In linear measurement models~\cite{candes2006robust, donoho2006compressed, foucart2013invitation}, the goal is to recover a high-dimensional signal $\mathbf{x}^* \in \mathbb{R}^N$ from noisy linear measurements:
\begin{equation}
    \label{eq:linear_inverse}
    \mathbf{y} = \mathbf{A}\mathbf{x}^* + \bm{\epsilon},
\end{equation}
where $\mathbf{A} = \left[ \mathbf a_1,\dots,\mathbf a_M\right]^\top\in \mathbb{R}^{M \times N}$ is the forward matrix, $\bm{\epsilon}=[\epsilon_1,\ldots,\epsilon_M]^\top \in \mathbb{R}^{M}$ is the noise vector, and $\mathbf{y} =[y_1,\ldots,y_M]^\top \in \mathbb{R}^{M}$ represents the acquired measurements. To address this typically ill-posed measurement model, especially in the high-dimensional regime where $M \ll N$, a series of works has developed effective recovery methods by exploiting hand-crafted structural priors of $\mathbf{x}^*$, such as sparsity or low-rankness~\cite{chen2013low, li2013compressed, nguyen2012robust, nguyen2013exact, xu2012outlier}.

However, the assumption of infinite measurement precision in linear measurement models is often unrealistic in hardware implementation. In practice, measurements are quantized to a finite number of bits. In particular, 1-bit quantization is valued for its low hardware cost and robustness to non-linear distortions~\cite{boufounos20081,boufounos2010reconstruction}. Both 1-bit compressed sensing (CS) and logistic regression are examples of the 1-bit quantized measurement model. For 1-bit CS, measurements are quantized to their signs:
\begin{equation}
    \label{eq:1bit_cs}
    \mathbf{y} = \text{sign}(\mathbf{A}\mathbf{x}^* + \bm\epsilon).
\end{equation}
For logistic regression, the likelihood of each binary outcome $y_i \in \{-1, +1\}$ can be expressed as follows:
\begin{equation}
\label{eq:logistic_background}
\mathbb{P}(y_i=1) = \frac{1}{1+e^{-\mathbf{a}_i^\top\mathbf{x}^*}}  = \frac{1}{2} + \frac{1}{2} \tanh\left( \frac{\mathbf{a}_i^\top \mathbf{x}}{2} \right).
\end{equation} 
1-bit quantized measurement models have also been widely studied under hand-crafted structural priors such as sparsity and low-rankness~\cite{plan2012robust,yan2012robust,plan2013one,jacques2013robust,davenport20141,knudson2016one,plan2017high}.

Priors have shifted from hand-crafted designs to leveraging the expressive capabilities of conventional generative models, such as variational autoencoders (VAEs) and generative adversarial networks (GANs)~\cite{bora2017compressed,liu2020information}. Recently, diffusion models (DMs)~\cite{sohl2015deep, ho2020denoising} stand out as powerful generative priors for capturing the statistical properties of natural images. Reconstruction methods using DMs generally fall into two categories: Supervised methods, which train neural networks end-to-end on paired clean and degraded images for specific measurement models, and unsupervised methods, which utilize pretrained unconditional diffusion models and knowledge of the degradation process. Recently, unsupervised methods have been favored for their versatility. They typically integrate a data-fidelity term with a DM-based prior within a gradient based optimization or sampling framework, achieving reliable reconstruction quality across a wide range of linear and non-linear measurement models~\cite{kadkhodaie2021stochastic, jalal2021robust, choi2021ilvr, chung2022improving, chung2023diffusion, zhu2023denoising, kawar2022denoising, wang2023zero, song2023pseudoinverse, song2024solving, mardani2024variational, zhang2025improving, zheng2025integrating}.

Specifically, under the assumption that the nonlinear forward operator $\mathcal{A}\,:\, \mathbb{R}^N \to \mathbb{R}^M$ is differentiable, approaches such as DPS~\cite{chung2023diffusion}, DiffPIR~\cite{zhu2023denoising}, and DAPS~\cite{zhang2025improving} can handle nonlinear measurement models of the form $\mathbf{y} = \mathcal{A}(\mathbf{x}^*) + \bm\epsilon$ through gradient based iterative procedures. However, these methods are not directly applicable to 1-bit CS or logistic regression, as the corresponding forward operators are either non-differentiable or lack an explicit formulation.

\textbf{Contributions:} To address this limitation, we draw inspiration from plug-and-play (PnP) frameworks~\cite{venkatakrishnan2013plug,chan2016plug,sreehari2016plug,kamilov2017plug} and propose Diff-OneBit, which is a PnP method designed to tackle the 1-bit CS problem using pretrained DMs. To overcome the challenge posed by the non-differentiable link function in 1-bit CS, our approach decouples the data and prior terms via half-quadratic splitting (HQS)~\cite{geman1995nonlinear} and introduces a differentiable surrogate function for the data-fidelity term, solving them iteratively within the diffusion sampling framework. We further adapt Diff-OneBit to address the logistic regression problem. Evaluations of Diff-OneBit on 1-bit CS and logistic regression tasks using the FFHQ and CelebA datasets demonstrate that it achieves superior image reconstruction quality with greater efficiency compared to competing methods.

\subsection{Related work}

In this subsection, we present an overview of relevant existing studies. These studies can broadly be categorized into (i) Signal recovery with conventional generative models, and (ii) Signal recovery with diffusion models.

\textbf{Signal recovery with conventional generative models:} Building on the pioneering work of~\cite{bora2017compressed}, prior research has explored signal recovery problems using conventional generative models. Subsequent studies have introduced novel architectures, including untrained deep image priors~\cite{van2018compressed}, underparameterized deep decoders~\cite{heckel2019deep}, and invertible networks designed to eliminate representation error~\cite{asim2020invertible}. Additionally, several methods with provable guarantees have been developed, such as those based on Langevin dynamics~\cite{nguyen2021provable} and posterior sampling~\cite{jalal2021instance}. The application scope of these priors has expanded to address problems like phase retrieval~\cite{hand2018phase,liu2021towards,liu2022misspecified}, generative principal component analysis~\cite{liu2022generative,chen2025solving} and generalized eigenvalue problems~\cite{liu2024generalized}. This progress has been supported by extensive studies confirming the adversarial robustness of these methods~\cite{genzel2022solving} and comprehensive surveys of the field~\cite{ongie2020deep,scarlett2023theoretical}.

Notably, there have been specific investigations into 1-bit CS using conventional generative models. For instance, the work~\cite{qiu2020robust} employ ReLU generative networks with dithering techniques to enable signal recovery. The work~\cite{liu2020sample} establish recovery guarantees for 1-bit CS using conventional generative models. Furthermore, single-index models encompassing 1-bit quantized measurement models have been studied under conventional generative priors~\cite{wei2019statistical,liu2020generalized,liu2022projected,liu2022non,chen2023unified}. However, these works focus on recovery guarantees related to optimal solutions for empirical risk minimization problems or approaches that approximately optimize over the range of conventional single-step generative models, without considering plug-and-play frameworks.

\textbf{Signal recovery with diffusion models:} A family of unsupervised methods using pretrained DMs excels at solving signal recovery problems by adapting the reverse diffusion process. For example, Score-ALD~\cite{jalal2021robust} utilizes Langevin dynamics for posterior sampling, while MCG~\cite{chung2022improving} adds manifold constraints in reverse sampling. ILVR~\cite{choi2021ilvr} refines each step based on reference images, and DDRM~\cite{kawar2022denoising} operates in the spectral domain of the forward operator. DiffPIR~\cite{zhu2023denoising} integrates plug-and-play frameworks. DPS~\cite{chung2023diffusion} extends posterior sampling to noisy and non-linear problems. DDNM~\cite{wang2023zero} introduces null-space guidance and $\Pi$GDM~\cite{song2023pseudoinverse} introduces pseudoinverse guidance in sampling. ReSample~\cite{song2024solving} utilizes hard data consistency for latent diffusion models. RED-diff~\cite{mardani2024variational} offers a variational perspective connecting diffusion to regularization by denoising. DAPS~\cite{zhang2025improving} employs a decoupled noise annealing process. However, the aforementioned methods are primarily designed for linear measurement models, while methods capable of handling non-linear measurement models, such as DPS, DiffPIR, DAPS, RED-diff, and $\Pi$GDM, are mainly designed for explicit and differentiable forward operators and cannot effectively handle quantized measurements. In the following, we discuss recent DM-based methods aiming for quantized signal recovery.

In QCS-SGM~\cite{meng2023quantized}, a DM serves as a prior for recovering signals from noisy quantized measurements through posterior sampling, integrating a noise-perturbed pseudo-likelihood score with the score of the DM. However, QCS-SGM relies on annealed Langevin dynamics, which demands a large number of iterations and incurs substantial computational overhead. SIM-DMIS~\cite{tang2025learning} investigates single index models that encompass 1-bit quantized measurement models with a diffusion prior. The approach first obtains an appropriate initial estimate, then reconstructs the underlying signal using a single round of unconditional sampling combined with partial inversion of the DM, which is an operation that approximates projecting the initial estimate onto the range the DM.

\section{Background}
\label{sec:background}
\subsection{Diffusion models}

DMs are generative models that map data to noise through a pre-defined forward diffusion process~\cite{ho2020denoising, karras2022elucidating,song2019generative, song2020improved,song2021score}. Specifically, the forward process perturbs a data sample $\mathbf{x}_0 \sim p_{\text{data}}$ over a continuous time variable $t \in [0, T]$ according to a Gaussian transition kernel:
\begin{equation}
    q_{0t}(\mathbf{x}_t | \mathbf{x}_0) = \mathcal{N}(\mathbf{x}_t; \alpha_t \mathbf{x}_0, \sigma_t^2 \mathbf{I}),
    \label{eq:forward_process}
\end{equation}
where $\alpha_t$ and $\sigma_t$ are time-dependent functions controlling the signal and noise schedules. Let $p_t$ denotes the marginal distribution of $\mathbf{x}_t$. As $t \to T$, the perturbed distribution $p_t(\mathbf{x}_t)$ converges to an isotropic Gaussian.

To generate a sample, DMs learn to reverse this process, starting from pure Gaussian noise $\mathbf{x}_T \sim \mathcal{N}(\bm 0,\mathbf{I})$ and iteratively denoising it to produce a sample from the data distribution. Specifically, a neural network $\bm\epsilon_\mathbf{\btheta}(\mathbf{x}_t,t)$ is trained to predict the noise component $\bm\epsilon$ from the noisy state $\mathbf{x}_t = \alpha_t \mathbf{x}_0 + \sigma_t \bm \epsilon$ by minimizing the following objective~\cite{song2019generative, song2021denoising}:
\begin{equation}
    \mathbb{E}_{\mathbf{x}_0\sim p_0, \bm\epsilon \sim \mathcal N(\bm0,\bm I), t \sim \mathcal U(0,T)} [\|\bm\epsilon_\mathbf{\btheta}(\mathbf{x}_t,t)-\bm\epsilon \|_2^2].
    \label{eq:dpm_denoising_objective}
\end{equation}
The pretrained noise predictor $\bm\epsilon_{\btheta}(\mathbf{x}_t, t)$ provides an estimate of the score function via the relation $\nabla_{\mathbf{x}_t} \log p_t(\mathbf{x}_t) \approx -\bm{\epsilon}_{\btheta}(\mathbf{x}_t, t) / \sigma_t$. Thus, sampling from DMs can be performed by numerically solving a stochastic differential equation (SDE)~\cite{song2021score}:
\begin{equation}
    \label{eq:sampling_sde}
    \mathrm{d}\mathbf{x}_t = \left( f(t) \mathbf{x}_t + \frac{g^2(t)}{\sigma_t} \bm{\epsilon}_{\btheta}(\mathbf{x}_t, t) \right) \mathrm{d}t + g(t) \mathrm{d}\mathbf{w}_t,
\end{equation}
where $\mathbf{w}_t$ represents a standard Wiener process, and $f(t)$ and $g(t)$ are the drift and diffusion coefficients, respectively. In Variance Preserving (VP) SDE, we have $f(t) = {\mathrm{d}\log\alpha_t} /{\mathrm{d}t}$, $g^2(t) = {\mathrm{d}\sigma_t^2}/{\mathrm{d}t} - 2 f(t) \sigma_t^2$. Alternatively, sampling can be performed deterministically by numerically solving the corresponding probability flow ordinary differential equation (ODE):
\begin{equation}
   \frac{\mathrm{d}\mathbf{x}_t}{\mathrm{d}t} = f(t) \mathbf{x}_t + \frac{g^2(t)}{2 \sigma_t} \bm{\epsilon}_{\btheta}(\mathbf{x}_t, t).
\label{eq:sampling_ode}
\end{equation}

\begin{figure*}[htbp]
    \centering
    \small
    \includegraphics[width=0.82\linewidth]{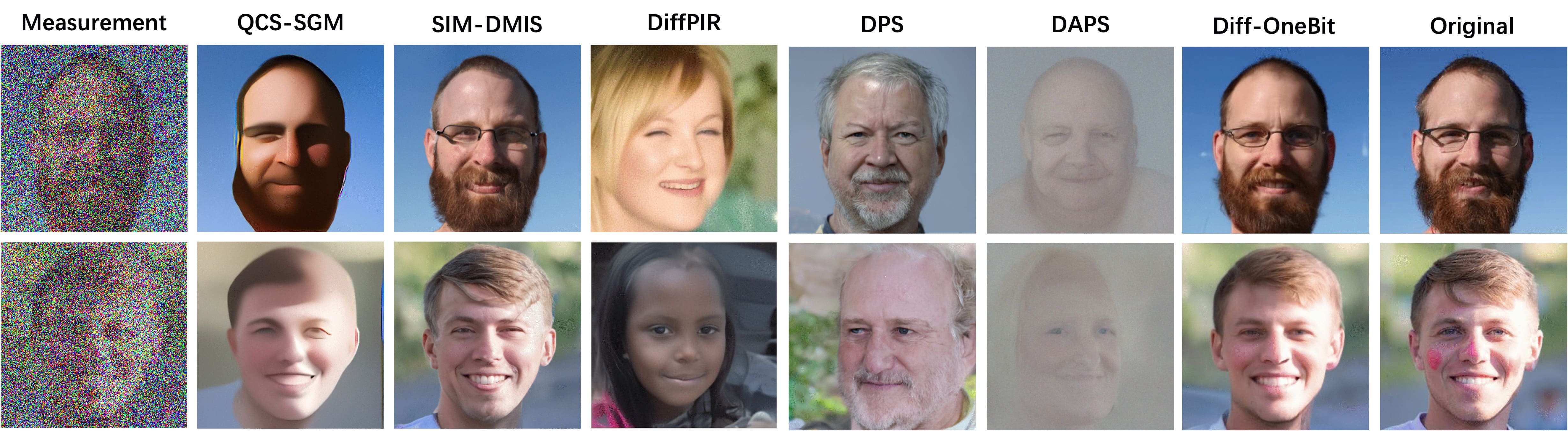}
    \caption{Qualitative results of 1-bit CS on FFHQ images. We compare Diff-OneBit with QCS-SGM, SIM-DMIS and DiffPIR. Input images have an addtive Gaussian noise of $\sigma=0.5$.}
    \label{fig:ffhq_visual_1bit_cs}
\end{figure*}

\subsection{Conditional diffusion models for signal recovery}

For signal recovery with measurements $\mathbf{y}$, the objective is to sample the clean signal $\mathbf{x}_0$ from the posterior distribution $p(\mathbf{x}_0|\mathbf{y})$. DMs achieve this by conditioning the reverse sampling process, modifying Eq.~\eqref{eq:sampling_ode} with the posterior score $\nabla_{\mathbf{x}_t} \log p_t(\mathbf{x}_t|\mathbf{y})$:
\begin{equation}
\label{eq:conditional_sde}
\frac{\mathrm{d}\mathbf{x}_t}{\mathrm{d}t} = f(t) \mathbf{x}_t - \frac{g^2(t)}{2} \nabla_{\mathbf{x}_t} \log p_t(\mathbf{x}_t|\mathbf{y}).
\end{equation}

For a linear measurement model with additive Gaussian noise $\bm\epsilon \sim \mathcal{N}(\mathbf{0}, \sigma^2 \mathbf{I})$ as in Eq.~\eqref{eq:linear_inverse}, the conditional score is directly related to the conditional expectation as follows~\cite{daras2024survey}:
\begin{equation}
\nabla_{\mathbf{x}_t} \log p_t(\mathbf{x}_t|\mathbf{y}) = \frac{\mathbb{E}[\mathbf{x}_0|\mathbf{x}_t,\mathbf{y}] - \mathbf{x}_t}{\sigma^2}.
\end{equation}
DiffPIR approximates this conditional expectation using HQS, which decouples the data-fidelity and prior terms as follows:
\begin{equation}
\mathbb{E}[\mathbf{x}_0|\mathbf{x}_t,\mathbf{y}] \approx \arg\min_{\mathbf{x}} \frac{1}{2} \|\mathbf{y} - \mathbf{A}\mathbf{x}\|_2^2 + \frac{\mu_t}{2} \|\mathbf{x} - \mathbb{E}[\mathbf{x}_0|\mathbf{x}_t]\|_2^2.
\end{equation}
Here, $\mu_t:=\lambda(\alpha_t^2 / \sigma_t^2)$ with $\lambda>0$ being a hyperparameter and $\mathbb{E}[\mathbf{x}_0|\mathbf{x}_t]$ can be directly obtained using Tweedie's formula~\cite{daras2024survey}.
Alternatively, approaches such as DPS approximate $\nabla_{\mathbf{x}_t} \log p_t(\mathbf{y}|\mathbf{x}_t)$ using Bayes' rule with 
\begin{equation}
    \nabla_{\mathbf{x}_t} \log p_t(\mathbf{x}_t|\mathbf{y}) = \nabla_{\mathbf{x}_t} \log p_t(\mathbf{y}|\mathbf{x}_t) + \nabla_{\mathbf{x}_t} \log p_t(\mathbf{x}_t),
\end{equation}
where the term $\nabla_{\mathbf{x}_t} \log p_t(\mathbf{x}_t)$ is simply the unconditional score function. However, these frameworks cannot be directly extended to handle 1-bit quantized measurement models for which both $\mathbb{E}[\mathbf{x}_0|\mathbf{x}_t,\mathbf{y}]$ and $\nabla_{\mathbf{x}_t} \log p_t(\mathbf{y}|\mathbf{x}_t)$ are difficult to approximate. 

\section{Method}
\label{sec:method}

Our goal is to address 1-bit quantized measurement models using a DM as a powerful generative prior. We formulate the reconstruction as a maximum a posterior (MAP) estimation problem. Given the 1-bit measurements $\mathbf{y}$, we seek to find the optimal vector $\hat{\mathbf{x}}$ that maximizes the posterior likelihood:
\begin{align}
    \hat{\mathbf{x}} &= \arg\max_{\mathbf{x}} p(\mathbf{x} | \mathbf{y}) = \arg\max_{\mathbf{x}} p(\mathbf{y} | \mathbf{x}) p(\mathbf{x}) \\
    &= \arg\min_{\mathbf{x}}  \underbrace{-\log p(\mathbf{y} | \mathbf{x})}_{\text{Data Fidelity Term}}\quad + \underbrace{(-\log p(\mathbf{x})).}_{\text{Prior Regularization Term}}
    \label{eq:map_objective}
\end{align}
In the following, we discuss the approaches designed for 1-bit CS and logistic regression, which are two popular instances of 1-bit quantization problems.

\subsection{Approach for 1-bit CS} %

For the 1-bit CS model in Eq.~\eqref{eq:1bit_cs}, we begin by modeling the underlying measurement process before quantization. For any $\mathbf{x} \in \mathbb{R}^n$, letting $\tilde{b}_i = \mathbf{a}_i^\top  \mathbf{x} + \epsilon_i$ be the corrupted measurement. The 1-bit measurement $y_i \in \{-1, +1\}$ is the result of applying the sign function to this noisy value:
\begin{equation}
    y_i = \text{sign}(\tilde{b}_i).
\end{equation}

The probability of observing $y_i = +1$ is the probability that the noisy measurement is non-negative:
\begin{align}
    \mathbb{P}(y_i =+1|\mathbf{x}) &= \mathbb{P}(\tilde{b}_i \ge 0)
    = \mathbb{P}(\mathbf{a}_i^\top \mathbf{x} + \epsilon_i \ge 0) \\
    &= \mathbb{P}(\epsilon_i \ge -\mathbf{a}_i^\top \mathbf{x}).\label{eq:prob_1}
\end{align}
Similarly, the probability of observing $y_i = -1$ is:
\begin{align}
    \mathbb{P}(y_i = -1|\mathbf{x}) &= \mathbb{P}(\tilde{b}_i < 0) = \mathbb{P}(\mathbf{a}_i^\top \mathbf{x} + \epsilon_i < 0) \\
    &= \mathbb{P}(\epsilon_i < -\mathbf{a}_i^\top \mathbf{x}).\label{eq:prob_minus_1}
\end{align}

Eqs.~\eqref{eq:prob_1} and~\eqref{eq:prob_minus_1} can be expressed using the cumulative distribution function (CDF). Assuming that the additive noise is Gaussian with $\epsilon_i \sim \mathcal{N}(0, \sigma^2)$, then $\epsilon_i / \sigma$ follows the standard normal distribution, with its CDF defined as $\Phi(z) = \int_{-\infty}^z \frac{1}{\sqrt{2\pi}} e^{-u^2/2} \mathrm{d}u$, and we obtain
\begin{align}
    \mathbb{P}(y_i = 1|\mathbf{x}) &= \mathbb{P}\left(\frac{\epsilon_i}{\sigma} \ge -\frac{\mathbf{a}_i^\top \mathbf{x}}{\sigma}\right) = \Phi\left( {\mathbf{a}_i^\top \mathbf{x} / {\sigma}} \right).
\end{align}
Applying the symmetry of the Gaussian distribution, we obtain $\mathbb{P}(\epsilon_i \ge v) = 1 - \mathbb{P}(\epsilon_i < v) = P(\epsilon_i \le -v)$, and thus
\begin{align}
    \mathbb{P}(y_i = -1|\mathbf{x}) &= \mathbb{P}\left(\frac{\epsilon_i}{\sigma} < -\frac{\mathbf{a}_i^\top \mathbf{x}}{\sigma}\right) = \Phi\left( -{\mathbf{a}_i^\top \mathbf{x} / {\sigma}} \right).
\end{align}
Combining these two cases into a single one, yielding
\begin{equation}
    \mathbb{P}(y_i|\mathbf{x}) = \Phi\left( {\mathbf{a}_i^\top \mathbf{x} / {\sigma}} \right)^{\frac{1+y_i}{2}} \cdot \Phi\left( -{\mathbf{a}_i^\top \mathbf{x} / {\sigma}} \right)^{\frac{1-y_i}{2}}.
\end{equation}
Then, we have the following for $\mathbf{y}$:
\begin{equation}
    \mathbb{P}(\mathbf{y}|\mathbf{x}) = \prod_{i=1}^M \Phi\left( {\mathbf{a}_i^\top \mathbf{x} / {\sigma}} \right)^{\frac{1+y_i}{2}} \cdot \Phi\left( -{\mathbf{a}_i^\top \mathbf{x} / {\sigma}} \right)^{\frac{1-y_i}{2}}.
    \label{eq:full_likelihood_product}
\end{equation}
This expression serves as our differentiable surrogate likelihood. For the MAP objective, we require its negative logarithm. Taking the negative logarithm of Eq.~\eqref{eq:full_likelihood_product} yields our final data-fidelity function as follows:
\begin{align}
   & \mathcal L(\mathbf{x};\mathbf{y}) = -\log p(\mathbf{y}|\mathbf{x}) \\&= -\sum_{i=1}^M \left[ \frac{1+y_i}{2} \log \Phi\left( \frac{\mathbf{a}_i^\top \mathbf{x}}{\sigma} \right) + \frac{1-y_i}{2} \log \Phi\left( -\frac{\mathbf{a}_i^\top \mathbf{x}}{\sigma} \right) \right]. \label{eq:surrogate_likelihood}
\end{align}

With a tractable data-fidelity term, we now address the MAP objective in Eq.~\eqref{eq:map_objective}. HQS introduces an auxiliary variable \(\mathbf{z}\) to decouple the data-fidelity and prior terms, reformulating the objective as
\begin{equation}
    \arg\min_{\mathbf{x}, \mathbf{z}} \mathcal{L}(\mathbf{x}; \mathbf{y}) - \log p(\mathbf{z}) \quad \text{subject to} \quad \mathbf{x} = \mathbf{z}.
\end{equation}

This constrained optimization problem is typically addressed by iteratively updating \(\mathbf{x}\) and \(\mathbf{z}\) through two subproblems, synchronized with the timestep \( t \) to ensure consistency with its reverse process schedule.

The \(\mathbf{x}\)-update rule enforces consistency with the observed 1-bit measurements using a differentiable surrogate likelihood. The subproblem is of the following form:
\begin{equation}
    \hat{\mathbf{x}} = \arg\min_{\mathbf{x}} \left( \mathcal{L}(\mathbf{x}; \mathbf{y}) + \frac{\mu}{2} \|\mathbf{x} - \hat{\mathbf{z}}\|_2^2 \right),
    \label{eq:x_update_problem}
\end{equation}
where \(\mu > 0\) is the penalty parameter and $\hat{\mathbf{z}}$ is an estimated vector. The differentiability of \(\mathcal{L}(\cdot;\mathbf{y})\) enables efficient optimization using gradient based methods such as stochastic gradient descent or Adam optimizer~\cite{kingma2014adam}. The \(\mathbf{z}\)-update rule involves solving an optimization problem of the following form:
\begin{equation}
    \hat{\mathbf{z}} = \arg\min_{\mathbf{z}} \left( -\log p(\mathbf{z}) + \frac{\mu}{2} \|\mathbf{z} - \hat{\mathbf{x}}\|_2^2 \right).\label{eq:DM-denoise}
\end{equation}
The optimization problem in Eq.~\eqref{eq:DM-denoise} can be thought of as a denoising problem, and its optimal solution can be approximated using a DM through the single-step Tweedie's formula, see, e.g., ~\cite{zhu2023denoising, venkatakrishnan2013plug}, for which we provide a detailed analysis in the appendix.

\begin{figure}[htbp]
    \small
    \centering
    \includegraphics[width=1\linewidth]{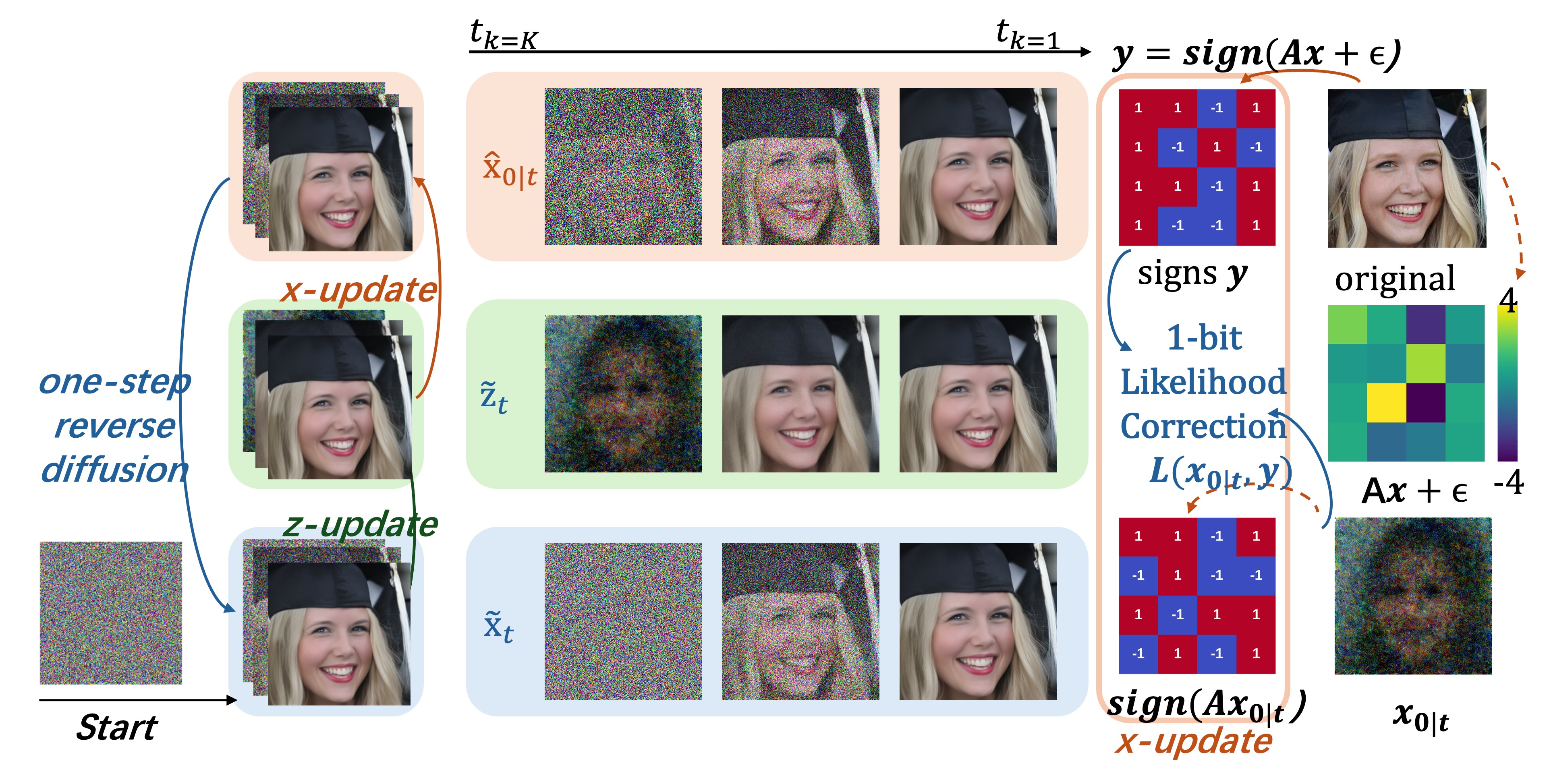}
    \caption{An illustration of the Diff-OneBit sampling step. At each reverse diffusion time $t$, we start with a noisy sample $\mathbf{\tilde x}_t$. First, the diffusion model acts as a denoiser to predict a preliminary clean image $\mathbf{\tilde z}_t$ (the prior update). This estimate is then corrected to a guided version $\hat{\mathbf{x}}_{0|t}$ by solving a data-consistency sub-problem using our differentiable surrogate likelihood. Finally, this guided estimate is used to compute the next state $\mathbf{\tilde x}_{t-1}$, advancing one step in the guided reverse diffusion.}
    \label{fig:illustration}
\end{figure}

Based on the above discussions, our Diff-OneBit approach is illustrated in Figure~\ref{fig:illustration} and detailed in Algorithm~\ref{alg:diff_onebit}.

\begin{figure*}[htbp]
    \centering
    \small
    \includegraphics[width=0.82\linewidth]{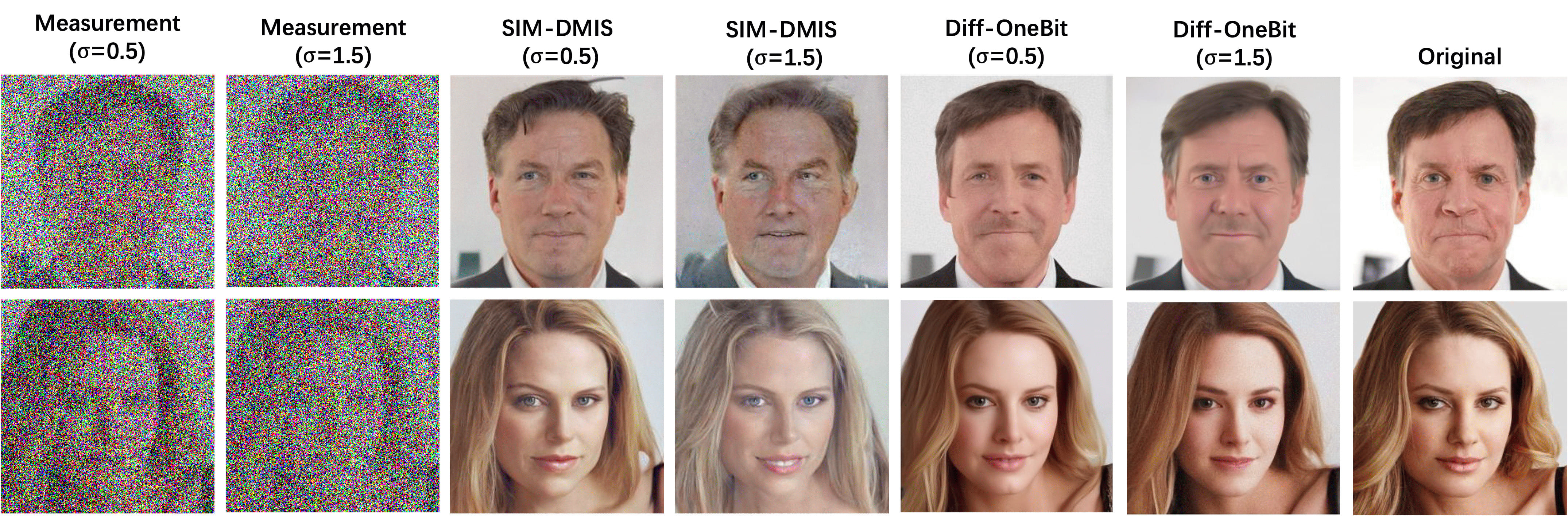}
    \caption{Qualitative results of 1-bit CS on CelebA images. We compare Diff-OneBit with SIM-DMIS. Input images have an additive Gaussian noise of $\sigma=0.5$ and $\sigma=1.5$.}
    \label{fig:celeba_visual_1bit_cs}
\end{figure*}

\subsection{Adaption for logistic regression}
We can adapt our Diff-OneBit approach designed for 1-bit CS to logistic regression. For the logistic regression model with 1-bit observation $y_i \in \{-1, +1\}$, the exact likelihood of an observation can be directly expressed using the hyperbolic tangent function. The probabilities for the two possible outcomes are:
\begin{align}
    \mathbb{P}(y_i = +1|\mathbf{x}) &= \frac{1}{2} + \frac{1}{2} \tanh\left( \frac{\mathbf{a}_i^\top \mathbf{x}}{2} \right), \\
    \mathbb{P}(y_i = -1|\mathbf{x}) &=  \frac{1}{2} - \frac{1}{2} \tanh\left( \frac{\mathbf{a}_i^\top \mathbf{x}}{2} \right).
\end{align}
Following the structure of the 1-bit model, we combine both cases into a single likelihood expression for $ \mathbb{P}(y_i|\mathbf{x})$. Analogous to the 1-bit CS formulation in Eq.~\eqref{eq:full_likelihood_product}, the exponents select the appropriate probability for the binary outcome:
\begin{equation}
    \mathbb{P}(\mathbf{y}|\mathbf{x}) = \prod_{i=1}^M  \mathbb{P}(y_i|\mathbf{x}).
    \label{eq:logistic_full_likelihood_product}
\end{equation}
Taking the negative logarithm yields the data-fidelity function, which corresponds to the standard binary cross-entropy loss:
\begin{align}
&\mathcal{L}(\mathbf{x}; \mathbf{y}) = -\log \mathbb{P}(\mathbf{y}|\mathbf{x}) \\
    &= -\sum_{i=1}^M \left( \frac{1+y_i}{2} \log \left( \frac{1}{2} + \frac{1}{2} \tanh\left( \frac{\mathbf{a}_i^\top \mathbf{x}}{2} \right) \right) + \frac{1-y_i}{2} \log \left( \frac{1}{2} - \frac{1}{2} \tanh\left( \frac{\mathbf{a}_i^\top \mathbf{x}}{2} \right) \right) \right). \label{eq:logistic_likelihood_final}
\end{align}

The function in Eq.~\eqref{eq:logistic_likelihood_final} is convex and smooth, making it well-suited for gradient based optimization within our MAP framework. Unlike the surrogate likelihood in Eq.~\eqref{eq:surrogate_likelihood}, this formulation represents the exact model likelihood and requires no hyperparameters such as the noise level $\sigma$.

\begin{algorithm}[htbp]
\caption{Diff-OneBit}
\label{alg:diff_onebit}
\begin{algorithmic}[1]
\Require Measurements $\mathbf{y}$, forward matrix $\mathbf{A}$, pretrained noise prediction network $\bm{\epsilon_\btheta}$, maximum number of iterations $K$, data-fidelity function $\mathcal L(\cdot;\mathbf y)$, schedules $\{\alpha_{t_k}, \sigma_{t_k}\}_{k=0}^K$, penalty coefficient $\lambda$, pre-calculated $\mu_{k} = \lambda(\alpha_{t_k}^2 / \sigma_{t_k}^2)$
\Ensure Reconstructed signal $\hat{\mathbf{x}}_0$
\State Sample $\mathbf{\tilde x}_{t_K} \sim \mathcal{N}(\mathbf{0}, \mathbf{I})$
\For{$k = K, \dots, 1$}
    \State $\mathbf{\tilde z}_{t_k} \leftarrow (\mathbf{\tilde x}_{t_k} - \sigma_{t_k} \bm{\epsilon}_{\btheta}(\mathbf{\tilde x}_{t_k}, t_k)) / \alpha_{t_k}$
    \State $\hat{\mathbf{x}}_{0|t_k} \leftarrow \arg\min_{\mathbf{x}} \left( \mathcal{L}(\mathbf{x}; \mathbf{y}) + \frac{\mu_{k}}{2} \|\mathbf{x} - \mathbf{\tilde z}_{t_k}\|_2^2 \right)$
    \State $\tilde{\bm{\epsilon}}_{t_k} \leftarrow (\mathbf{\tilde x}_{t_k} - \alpha_{t_k} \hat{\mathbf{x}}_{0|t_k}) / \sigma_{t_k}$
    \State $\mathbf{\tilde x}_{t_{k-1}} \leftarrow \alpha_{t_{k-1}} \hat{\mathbf{x}}_{0|t_k} + \sigma_{t_{k-1}} \tilde{\bm{\epsilon}}_{t_k}$
\EndFor
\State \Return $\mathbf{\tilde x}_{t_0}$
\end{algorithmic}
\end{algorithm}

\section{Experiments}
\label{sec:experiments}

In this section, we demonstrate the performance of our Diff-OneBit approach on 1-bit CS and logistic regression tasks. 

\textbf{Experimental setup.} Our experiments employ pretrained DMs that correspond to image datasets at a resolution of 256$\times$256. Specifically, for FFHQ dataset~\cite{karras2019style}, we utilized the score model~\cite{song2020improved} for QCS-SGM and the denoising model provided by~\cite{chung2023diffusion} for Diff-OneBit, DPS, DAPS, DiffPIR and SIM-DMIS. For CelebA~\cite{karras2018progressive} datasets, we used the models pretrained by~\cite{ho2020denoising}.\footnote{Since QCS-SGM does not provide a configuration file for reproducing results on the CelebA dataset, and we have not yet trained a score model on CelebA, we only tested the performance of QCS-SGM on the FFHQ dataset.} For ImageNet~\cite{deng2009imagenet}, we utilized the models pretrained by~\cite{dhariwal2021diffusion}. All the experiments are conducted on NVIDIA Tesla V100. 

\textbf{Measurement tasks:} In all experiments, we evaluate our method on the validation set of 100 images on 1-bit CS and logistic regression. We have a forward matrix with i.i.d. entries drawn from a normal distribution $\mathcal{N}(0, 1/M)$. Measurements are evaluated under Gaussian noises with a measurement ratio of $M/N = 1/16$. 

\textbf{Baselines and metrics.} We compare our proposed Diff-OneBit approach with several state-of-the-art methods. The primary baselines are specialized solvers for quantized measurements, including QCS-SGM and SIM-DMIS. Additionally, we include DiffPIR, DPS, and DAPS, that can handle non-linear measurement models, in our comparison. However, their non-linear frameworks are incompatible with 1-bit quantization, as the link function is non-differentiable or cannot be explicitly expressed. Thus, for our comparison, we use their linear versions, ignoring the non-linear link function. Reconstruction quality is evaluated using a comprehensive set of metrics. For fidelity and faithfulness to the ground truth, we use the peak signal-to-noise ratio (PSNR) and the structural similarity index measure (SSIM). For perceptual quality, we use the Learned Perceptual Image Patch Similarity (LPIPS)~\cite{zhang2018unreasonable} score and the Fr\'{e}chet Inception Distance (FID)~\cite{heusel2017gans}.

\begin{table}[htbp]
\centering
\begin{tabular}{l l c c c c c c c c}
\toprule
\textbf{FFHQ} & & \multicolumn{4}{c}{1-bit CS ($\sigma=0.5$)} & \multicolumn{4}{c}{1-bit CS ($\sigma=1.5$)} \\
\cmidrule(lr){3-6} \cmidrule(lr){7-10}
\textbf{Method} & NFE & PSNR$\uparrow$ & SSIM$\uparrow$ & LPIPS$\downarrow$ & FID$\downarrow$ & PSNR$\uparrow$ & SSIM$\uparrow$ & LPIPS$\downarrow$ & FID$\downarrow$ \\
\midrule
Diff-OneBit & 20 & $\textbf{22.05}\pm1.88$ & $\textbf{0.64}\pm.07$ & $\textbf{0.34}\pm.05$ & \textbf{85.40} & $\textbf{20.25}\pm1.01$ & $\textbf{0.53}\pm.06$ & $\textbf{0.44}\pm.05$ & \textbf{96.97} \\
SIM-DMIS & 150 & $18.97\pm2.36$ & $0.54\pm.09$ & $0.42\pm.05$ & 110.05  & $14.32\pm1.85$ & $0.41\pm.08$ & $0.56\pm.06$ & 159.38 \\
DiffPIR & 100 & $9.88\pm1.41$ & $0.13\pm.04$ & $0.78\pm.05$ & 219.24 & $9.79\pm1.39$ & $0.12\pm.04$ & $0.78\pm.05$ & 226.36 \\
DPS & 1000 & $12.37\pm1.70$ & $0.31\pm.08$ & $0.64\pm.07$ & 132.84  & $11.68\pm1.47$ & $0.29\pm.08$ & $0.65\pm.06$ & 133.02  \\
DAPS & 1000 & $11.30\pm1.60$ & $0.37\pm.08$ & $0.69\pm.05$ & 203.12 & $11.39\pm1.54$ & $0.34\pm.08$ & $0.73\pm.05$ & 224.04 \\
QCS-SGM & 11555 & $18.88\pm2.30$ & $0.52\pm.10$ & $0.55\pm.06$ & 211.85 & $16.84\pm0.98$ & $0.47\pm.08$ & $0.64\pm.06$ & 297.27 \\
\toprule
\textbf{CelebA} & & \multicolumn{4}{c}{1-bit CS ($\sigma=0.5$)} & \multicolumn{4}{c}{1-bit CS ($\sigma=1.5$)} \\
\cmidrule(lr){3-6} \cmidrule(lr){7-10}
\textbf{Method} & NFE & PSNR$\uparrow$ & SSIM$\uparrow$ & LPIPS$\downarrow$ & FID$\downarrow$ & PSNR$\uparrow$ & SSIM$\uparrow$ & LPIPS$\downarrow$ & FID$\downarrow$ \\
\midrule
Diff-OneBit & 20 & $\textbf{22.48}\pm1.80$ & $\textbf{0.66}\pm.09$ & $\textbf{0.31}\pm.05$  & \textbf{66.21} & $\textbf{20.75}\pm1.03$ & $\textbf{0.55}\pm.07$ & $\textbf{0.40}\pm.04$ & \textbf{71.81} \\
SIM-DMIS & 150 & $18.88\pm2.50$ & $0.56\pm.09$ & $0.38\pm.05$ & 75.93 & $14.78\pm2.36$ & $0.43\pm.09$ & $0.51\pm.06$ & 102.75 \\
DiffPIR & 100 & $9.20\pm1.51$ & $0.13\pm.05$ & $0.81\pm.10$ & 184.89  & $9.12\pm1.49$ & $0.12\pm.05$ & $0.82\pm.10$ & 185.24  \\
DPS & 1000 & $12.94\pm1.82$ & $0.36\pm.09$ & $0.56\pm.07$ & 97.01  & $11.78\pm1.71$ & $0.31\pm.08$ & $0.60\pm.06$ & 105.19  \\
DAPS & 1000 & $11.33\pm1.62$ & $0.38\pm.08$ & $0.67\pm.05$ & 186.51 & $11.32\pm1.53$ & $0.35\pm.08$ & $0.72\pm.05$ & 211.46 \\
\midrule
\textbf{ImageNet} & & \multicolumn{4}{c}{1-bit CS ($\sigma=0.5$)} & \multicolumn{4}{c}{1-bit CS ($\sigma=1.5$)} \\
\cmidrule(lr){3-6} \cmidrule(lr){7-10}
\textbf{Method} & NFE & PSNR$\uparrow$ & SSIM$\uparrow$ & LPIPS$\downarrow$ & FID$\downarrow$ & PSNR$\uparrow$ & SSIM$\uparrow$ & LPIPS$\downarrow$ & FID$\downarrow$ \\
\midrule
Diff-OneBit & 20 & $\textbf{20.56}\pm2.34$ & $\textbf{0.52}\pm.16$ & $\textbf{0.47}\pm.09$  & \textbf{92.71} & $\textbf{18.23}\pm1.72$ & $\textbf{0.43}\pm.15$ & $\textbf{0.55}\pm.07$ & \textbf{121.81} \\
SIM-DMIS & 150 & $17.37\pm2.68$ & $0.41\pm.15$ & $0.50\pm.08$ & {95.72} & $14.87\pm2.79$ & $0.30\pm.15$ & $0.61\pm.09$ & 138.37 \\
\bottomrule
\end{tabular}
\caption{Quantitative results of 1-bit CS on FFHQ, CelebA and ImageNet images. We compare our method with DiffPIR, DPS, DAPS QCS-SGM and SIM-DMIS. Input images have an additive Gaussian noise of $\sigma=0.5$ and $\sigma=1.5$.}
\label{tab:ffhq_celeba_1bit_tasks}
\end{table}

\begin{table}[htbp]
\centering
\small
\begin{tabular}{l c c c}
\toprule
 NFE & PSNR$\uparrow$ & SSIM$\uparrow$ & LPIPS$\downarrow$ \\
\midrule
20 & ${22.05} \pm 1.88$ & ${0.64} \pm .07$ & ${0.34} \pm .05$ \\
50 & ${22.39} \pm 1.34$ & ${0.65} \pm .06$ & ${0.33} \pm .05$ \\
100 & ${22.65} \pm 1.08$ & ${0.65} \pm .06$ & ${0.32} \pm .05$ \\
\bottomrule
\end{tabular}
\caption{Quantitative results of 1-bit CS on FFHQ images under different NFEs. We evaluate our method under 20, 50 and 100 NFEs with additive Gaussian noise of $\sigma=0.5$.}
\label{tab:NFE_tasks}
\end{table}

\subsection{Quantitative results}

For 1-bit CS, the main quantitative results for the FFHQ, CelebA and ImageNet datasets are presented in Table~\ref{tab:ffhq_celeba_1bit_tasks}. Our proposed method, Diff-OneBit, consistently outperforms all baseline methods across all evaluation metrics, demonstrating its superior reconstruction accuracy. For logistic regression, the main quantitative results on the FFHQ dataset are presented in Table~\ref{tab:ffhq_logistic_regression_tasks}.
We compare our method with DPS, DAPS, and DiffPIR, which are capable of handling non-linear measurements. We also compare with SIM-DMIS, as logistic regression is a special case of SIM.

Our experiments demonstrate that existing solvers for linear or non-linear measurement models perform poorly in 1-bit quantization tasks. Tables~\ref{tab:ffhq_celeba_1bit_tasks} and~\ref{tab:ffhq_logistic_regression_tasks} show that DiffPIR, DPS, and DAPS significantly underperforms compared to specialized methods and our proposed Diff-OneBit approach. These results demonstrate that optimization and gradient based methods, designed for linear or non-linear problems, are ineffective for 1-bit measurement models.

\begin{table}[htbp]
\centering
\small
\begin{tabular}{l c c c}
\toprule
\textbf{FFHQ} & \multicolumn{3}{c}{logistic regression}\\
\cmidrule(lr){2-4}
\textbf{Method} & PSNR$\uparrow$ & SSIM$\uparrow$ & LPIPS$\downarrow$ \\
\midrule
Diff-OneBit & $\textbf{19.99}\pm1.94$ & $\textbf{0.56}\pm.08$ & $\textbf{0.42}\pm.05$ \\
SIM-DMIS & $16.03\pm2.34$ & $0.46\pm.09$ & $0.49\pm.06$ \\
DiffPIR & $9.84\pm1.40$ & $0.13\pm.04$ & $0.79\pm.05$ \\
DPS & $15.03\pm2.37$ & $0.43\pm.11$ & $0.53\pm.08$  \\
DAPS  & $13.43\pm1.64$ & $0.40\pm.08$ & $0.63\pm.07$ \\
\bottomrule
\end{tabular}
\caption{Quantitative results of logistic regression on FFHQ images. We compare Diff-OneBit with DAPS, DPS, DiffPIR, and SIM-DMIS without additional Gaussian noise, setting NFE to 20 for Diff-OneBit, 100 for DiffPIR, 150 for SIM-DMIS, and 1000 for DPS and DAPS.}
\label{tab:ffhq_logistic_regression_tasks}
\end{table}

\subsection{Qualitative results}

\begin{figure}
    \centering
    \includegraphics[width=1\linewidth]{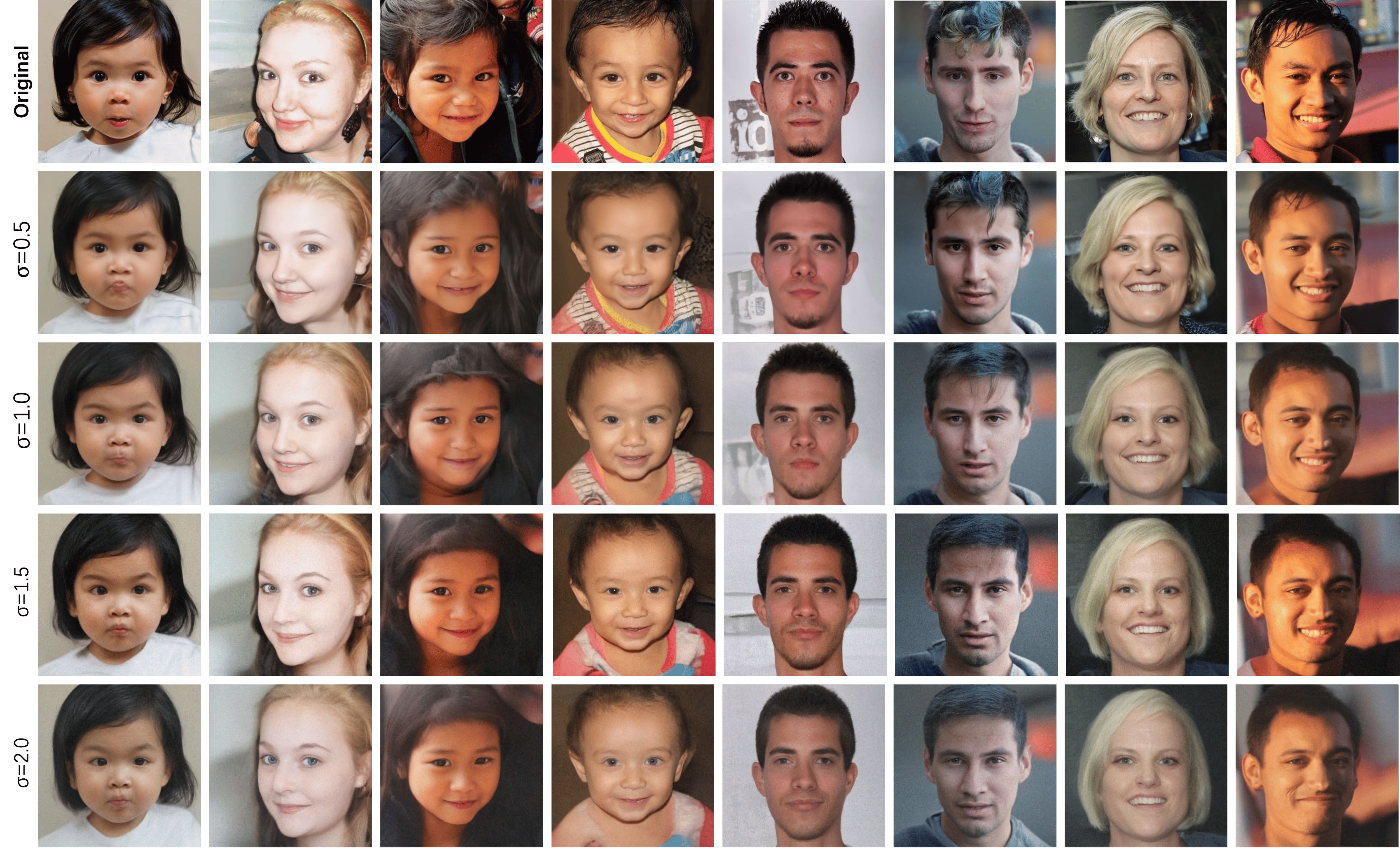}
    \caption{Qualitative results of 1-bit CS on FFHQ images. We compare Diff-OneBit under varying Gaussian noise levels in 1-bit CS.}
    \label{fig:ffhq_1bitCS_sigma}
\end{figure}

\begin{figure}
    \centering
    \includegraphics[width=1\linewidth]{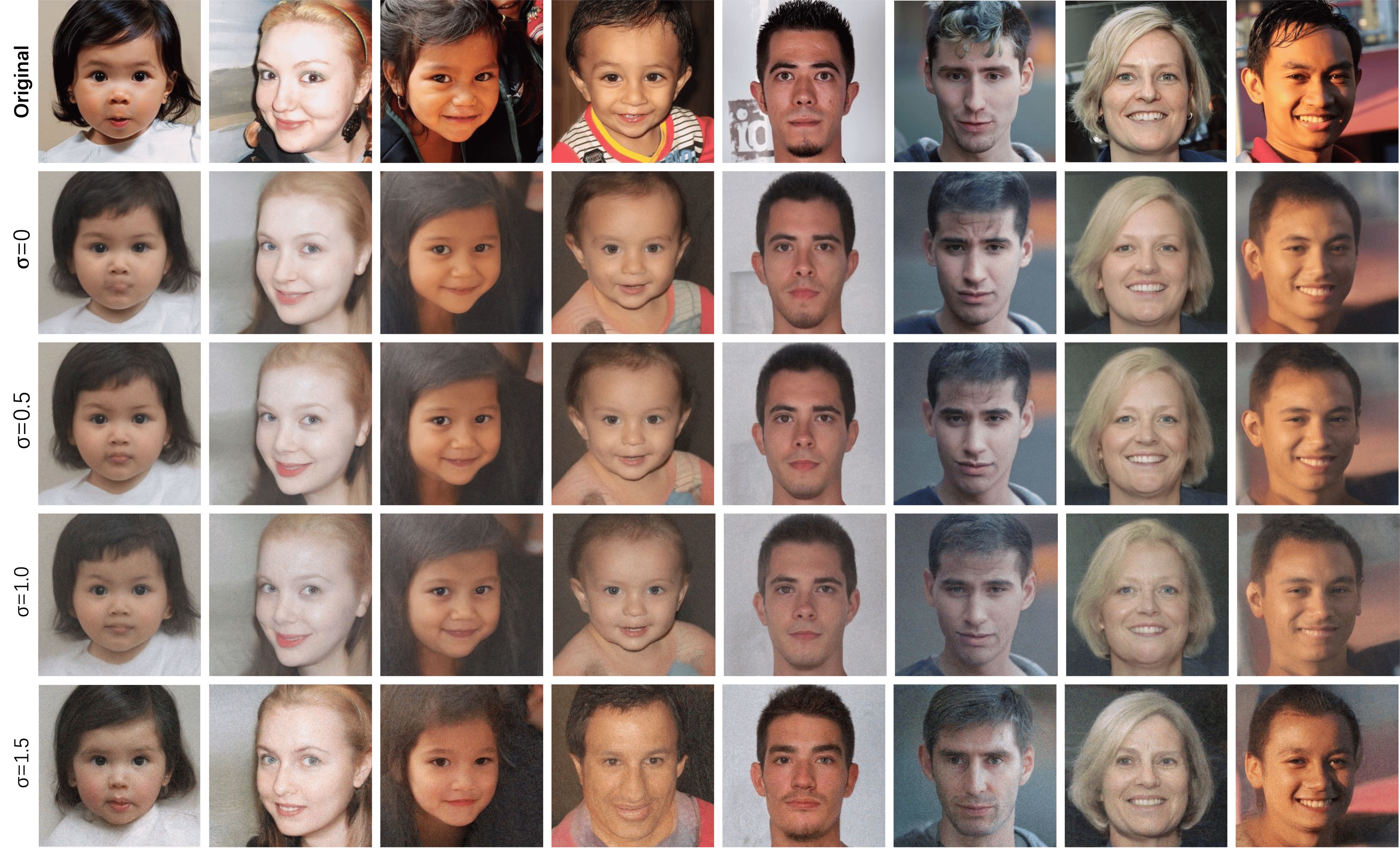}
    \caption{Qualitative results of logistic regression on FFHQ images. We compare Diff-OneBit under varying Gaussian noise levels in logistic regression.}
    \label{fig:ffhq_1bit_lr_sigma}
\end{figure}

\begin{figure}[htbp]
    \centering
    \includegraphics[width=1\linewidth]{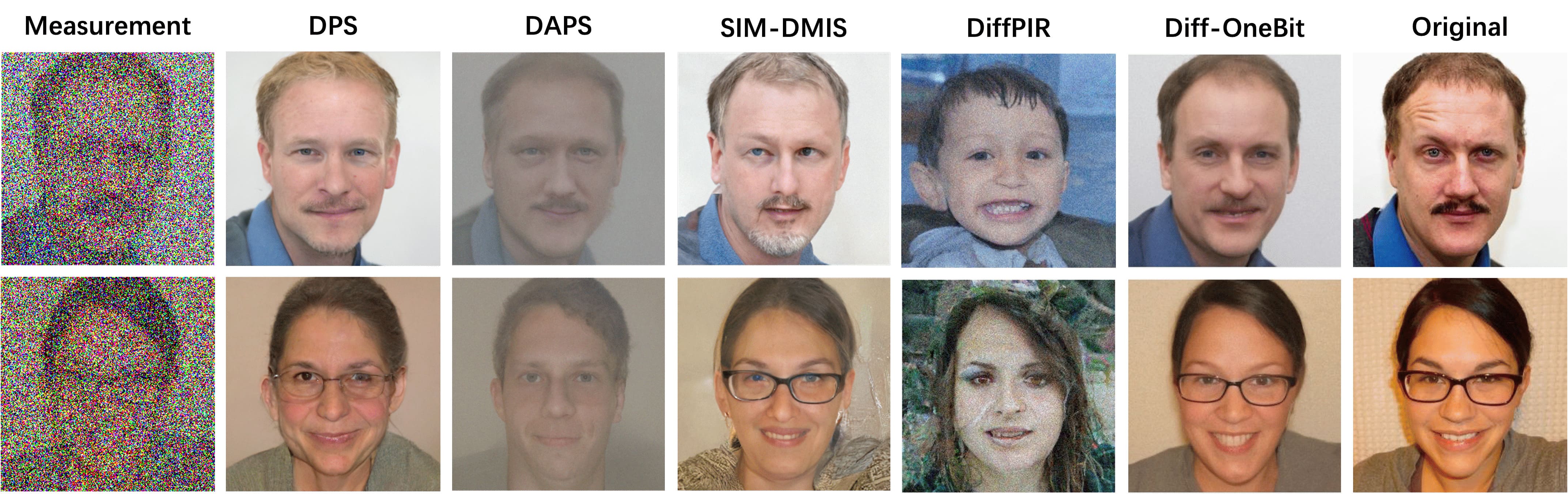}
    \caption{Qualitative results of logistic regression on FFHQ images. We compare Diff-OneBit with DPS, DAPS, DiffPIR and SIM-DMIS.}
    \label{fig:ffhq_visual_1bit_lr}
\end{figure}

For 1-bit CS, Figures~\ref{fig:ffhq_visual_1bit_cs} and~\ref{fig:celeba_visual_1bit_cs} provide visual comparison of the reconstruction results on FFHQ and CelebA images for the 1-bit CS.\footnote{We use $\mathbf{A}^\top\mathbf{y}$ to denote the measurement, ensuring dimensional consistency.} Images reconstructed by Diff-OneBit exhibit significantly fewer artifacts and preserve finer textural details and structural integrity. In contrast, specialized solvers like QCS-SGM produce reconstructions with noticeable smoothing, while SIM-DMIS results in loss of high-frequency details. DiffPIR, DPS, and DAPS fail to capture the semantic content of images from 1-bit measurements using gradient information. For logistic regression, Figure~\ref{fig:ffhq_visual_1bit_lr} provides visual comparison of the reconstruction results on FFHQ images for logistic regression. Images reconstructed by our Diff-OneBit approach exhibit significantly fewer artifacts and preserve finer textural details and structural integrity.\\
Moreover, Figures~\ref{fig:ffhq_1bitCS_sigma} and~\ref{fig:ffhq_1bit_lr_sigma} provide a visual comparison of reconstructed FFHQ images for 1-bit CS and logistic regression based on the Diff-OneBit method under varying Gaussian noise levels, specifically with $\sigma \in \{0.5, 1.0, 1.5, 2.0\}$ for 1-bit CS and $\sigma \in \{0.0, 0.5, 1.0, 1.5\}$ for logistic regression. These figures demonstrate that our approach consistently achieves high-fidelity image recovery across different noise intensities, showing robust performance in preserving details, for which detailed quantitative results are provided in the appendix.


\subsection{Ablation studies and further analysis}
To validate the design choices and flexibility of our method, we conduct further analysis.

\begin{figure}
    \centering
    \includegraphics[width=1\linewidth]{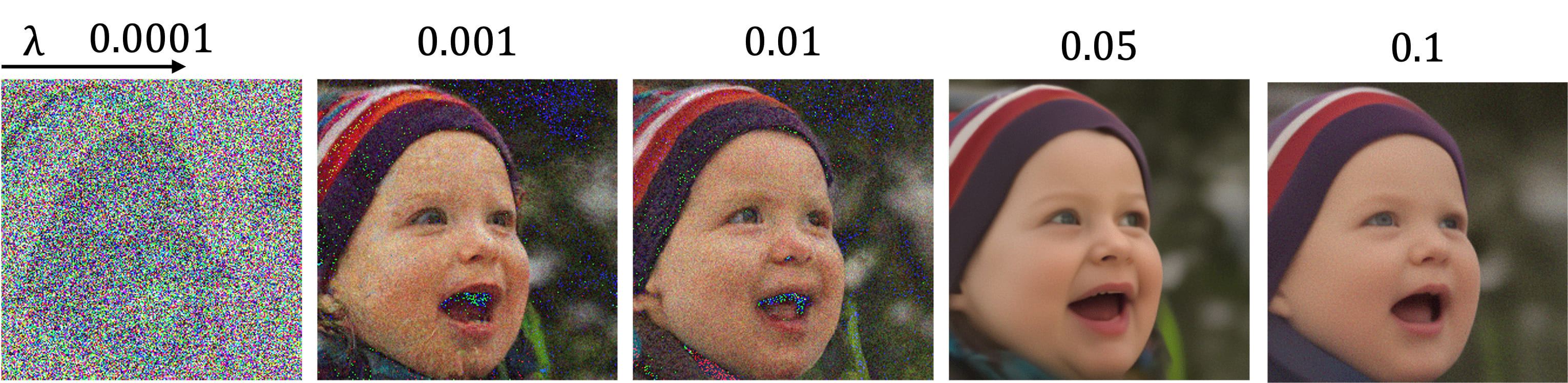}
    \caption{Effect of hyperparameters \(\lambda\). Reconstructed images from the same 1-bit CS measurement with input images corrupted by additive Gaussian noise of \(\sigma = 0.5\).}
    \label{fig:effect_ablation}
\end{figure}

\noindent\textbf{Experiment under varying NFEs.} We evaluate our method by varying NFEs from 20 to 100 for 1-bit CS on 100 FFHQ images with Gaussian noise \(\sigma=0.5\). Table~\ref{tab:NFE_tasks} demonstrates that our method is efficient even at low NFEs. Moreover, the reconstruction quality further improves as NFE increases from 20 to 100, showing the scalability of our method.

\noindent\textbf{Comparison with DiffPIR.} Our method, similar to DiffPIR, employs the HQS approach but differs in two major aspects. First, DiffPIR lacks specific modeling for 1-bit measurements, resulting in inferior reconstruction, as shown in our experiments. Then, unlike DiffPIR, which adds Gaussian noise during sampling, our algorithm performs better without random noise. The impact of varying random noise levels is presented in the appendix. We adopt similar hyperparameters, with \(\lambda\) controlling the data-fidelity term via \(\mu_t = \lambda ({\alpha_t^2}/{\sigma_t^2})\). Figure~\ref{fig:effect_ablation} demonstrates that our method, Diff-OneBit, exhibits robust performance across a range of \(\lambda\) values on the FFHQ dataset. However, overly strong (\(\lambda < 0.0001\)) or weak (\(\lambda \geq 0.1\)) guidance causes noisy or overly smooth reconstructions, respectively.

\noindent\textbf{Computational efficiency.}
Our algorithm achieves high-quality reconstructions using only 20 NFEs. However, the gradient descent optimization introduces additional computational overhead. To ensure a fair performance comparison, we provide a detailed analysis of the time overhead in the appendix.

\section{Conclusion}
\label{sec:conclusion}
In this paper, we address the challenge of signal recovery under 1-bit quantization in the context of 1-bit compressed sensing and logistic regression using diffusion models. We propose Diff-OneBit, an approach that integrates a differentiable surrogate likelihood function with a plug-and-play optimization scheme, leveraging a pretrained diffusion model. Experimental results demonstrate that Diff-OneBit achieves state-of-the-art reconstruction quality on the FFHQ and CelebA datasets, offering high computational efficiency and versatility across 1-bit quantized measurement models.

\appendices

\section{Experimental results on 1-bit random inpainting}

To further validate the effectiveness of our proposed method on 1-bit quantization tasks, we introduce 1-bit random inpainting. The degradation model for this masked image inpainting task can be expressed as
\begin{equation}
    \mathbf{y} = \text{sign}(\mathbf{P} \mathbf{x^*} + \mathbf{n}),
\end{equation}
where $\mathbf{P}$ is a user-defined random binary mask that corresponding to a subset of pixels from the original image $\mathbf{x^*}$, and $\mathbf{n}$ represents additive Gaussian noise. In this task, the degradation process involves two steps: First, the image $\mathbf{x}$ is partially observed through the mask $\mathbf{P}$, and then the observed pixels are subjected to additive Gaussian noise $\mathbf{n}$ before being quantized to their sign. The image inpainting task is to recover the missing and distorted information.

For our experiments, we use a 50\% mask ratio, meaning 50\% of the pixels are observed, with a noise level of $\sigma = 0.5$. This quantization introduces significant non-linearity, as a substantial amount of information from the original pixels is discarded.

\begin{figure}[htbp]
    \centering
    \includegraphics[width=0.8\linewidth]{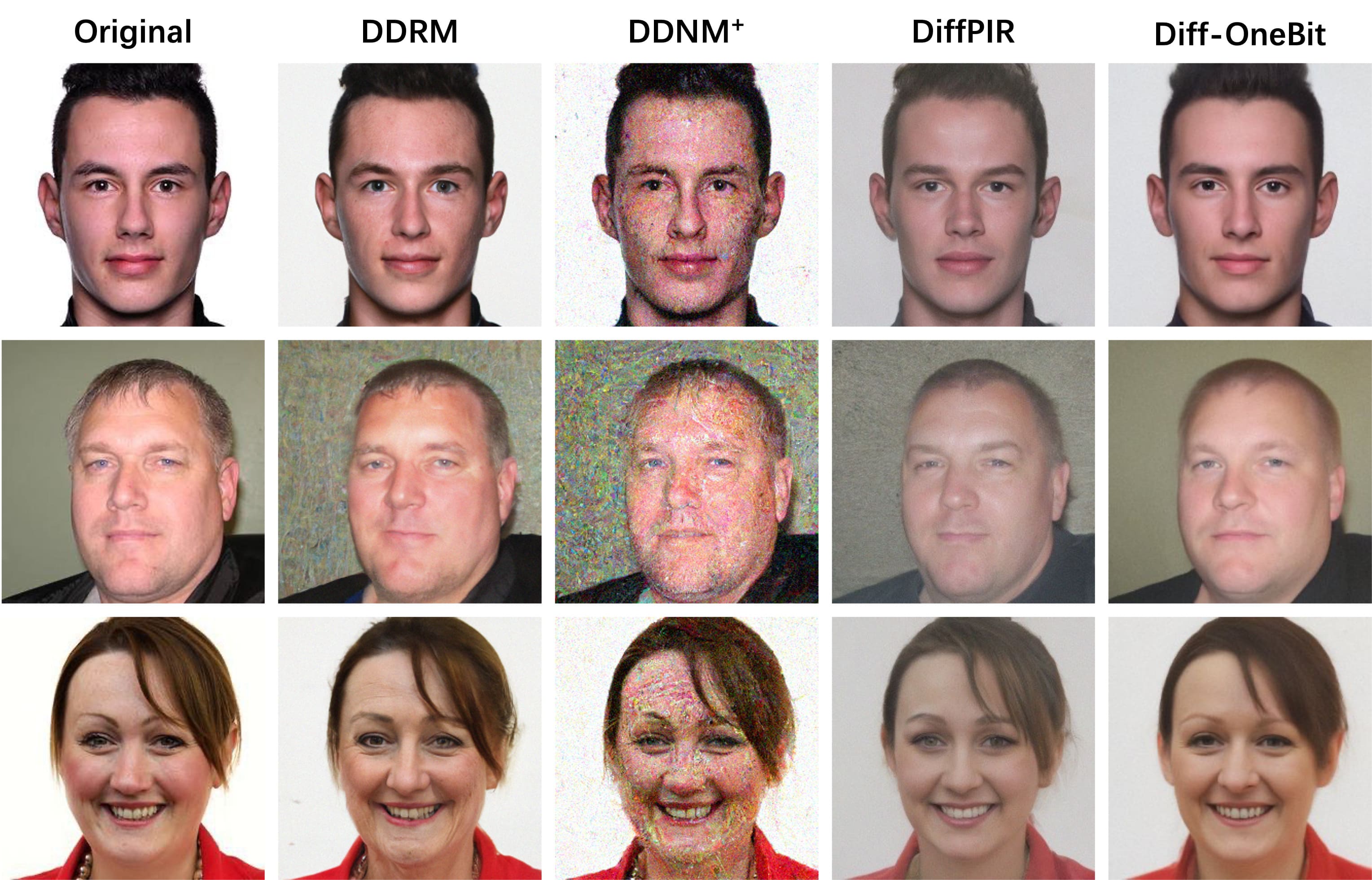}
    \caption{{Qualitative results of 1-bit random inpainting on FFHQ images.}}
    \label{fig:1bit_inpainting_results}
\end{figure}

We compare Diff-OneBit against DDRM, DDNM$^+$, and DiffPIR, which can handle noisy linear measurement models. For DDRM, DDNM$^+$, and DiffPIR, We use linear measurements which contain substantially more information than the 1-bit quantized measurements that our method uses. We test 100 images on FFHQ validation sets. The results summarized in Table~\ref{tab:inpainting_results} and visualized in Figure~\ref{fig:1bit_inpainting_results}, clearly demonstrate the superiority of our approach.

\begin{table}[htbp]
\centering
\begin{tabular}{l c c c}
\toprule
{Method} & {PSNR}$\uparrow$ & {SSIM}$\uparrow$ & {LPIPS}$\downarrow$ \\
\midrule
\multicolumn{4}{c}{$\mathbf{y} = \mathbf{Px^*} + \mathbf{n}$ (Linear Measurement)} \\
\midrule
DDRM & $23.09 \pm 1.03$ & $0.56 \pm 0.05$ & $0.40 \pm 0.04$ \\
DDNM$^+$ & $18.97 \pm 0.58$ & $0.25 \pm 0.04$ & $0.61 \pm 0.05$ \\
DiffPIR & $17.67 \pm 1.21$ & $0.52 \pm 0.07$ & $0.46 \pm 0.04$ \\
\midrule
\multicolumn{4}{c}{$\mathbf{y} = \text{sign}(\mathbf{Px^*} + \mathbf{n})$ (Quantized, 1-Bit Measurement)} \\
\midrule
{Diff-OneBit} & ${20.89 \pm 1.29}$ & ${0.61 \pm 0.08}$ & ${0.38 \pm 0.05}$ \\
\bottomrule
\end{tabular}
\caption{Quantitative results of 1-bit random inpainting on FFHQ images. We consider random inpainting with a 50\% mask ratio and an additive Gaussian noise $\sigma=0.5$. We compare Diff-OneBit with DDRM, DDNM$^+$ and DiffPIR, setting NFE=20 for all methods.}
\label{tab:inpainting_results}
\end{table}

\textbf{Quantitative analysis.} As shown in Table~\ref{tab:inpainting_results}, Diff-OneBit achieves significantly better quantitative metrics. It obtains a PSNR of $20.89$, an SSIM of $0.61$, and an LPIPS of $0.38$. Remarkably, our method outperforms DiffPIR and DDNM$^+$ even though it operates on severely degraded, 1-bit measurements. For example, the best-performing linear baseline, DDRM, achieves an LPIPS of $0.40$ with linear measurement, whereas our method achieves lower LPIPS with 1-bit measurement. This highlights the effectiveness of our proposed method on 1-bit inpainting random.

\textbf{Qualitative analysis.} The qualitative comparisons in Figure~\ref{fig:1bit_inpainting_results} further underscore the effectiveness of Diff-OneBit. The images reconstructed by Diff-OneBit from 1-bit measurements exhibit higher fidelity to the ground truth compared to those reconstructed by the baseline methods from linear measurements.

\section{The influence of the Gaussian noise in Diff-OneBit}

In this section, we analyze the effect of introducing random Gaussian noise at each timestep, a technique adopted from DiffPIR. Specifically, we follow the reverse process of DiffPIR by incorporating a weighting factor, $\zeta$, a scalar controlling the strength of stochasticity, to replace the deterministic noise component with stochastic Gaussian noise. Our algorithm is presented in Algorithm~\ref{alg:diff_onebit_random}. As demonstrated in DiffPIR, introducing stochasticity can lead to improved reconstruction quality for linear measurement models. However, for 1-bit quantization, we found that using purely deterministic noise (i.e., $\zeta=0$) in the reverse process yields superior reconstruction quality.

\begin{algorithm}[t]
\caption{Diff-OneBit (with random noise)}
\label{alg:diff_onebit_random}
\begin{algorithmic}[1]
\Require Measurements $\mathbf{y} \in \mathbb{R}^{M}$, degradation matrix $\mathbf{A} \in \mathbb{R}^{M\times N}$, pre-trained noise prediction network $\bm{\epsilon_\btheta}$, maximum number of iterations $K$, data-fidelity function $\mathcal L(\cdot;\mathbf y)$, schedules $\{\alpha_{t_k}, \sigma_{t_k}\}_{k=0}^K$, penalty coefficient $\lambda$, pre-calculated $\mu_{k} = \lambda(\alpha_{t_k}^2 / \sigma_{t_k}^2)$, stochasticity level \( \zeta \)
\Ensure Reconstructed signal $\hat{\mathbf{x}}_0$
\State Sample $\mathbf{\tilde x}_{t_K} \sim \mathcal{N}(\mathbf{0}, \mathbf{I}_N)$
\For{$k = K, \dots, 1$}
    \State $\mathbf{\tilde z}_{t_k} \leftarrow (\mathbf{\tilde x}_{t_k} - \sigma_{t_k} \bm{\epsilon}_{\btheta}(\mathbf{\tilde x}_{t_k}, t_k)) / \alpha_{t_k}$
    \State $\hat{\mathbf{x}}_{0|t_k} \leftarrow \arg\min_{\mathbf{x}} \left( \mathcal{L}(\mathbf{x}; \mathbf{y}) + \frac{\mu_{k}}{2} \|\mathbf{x} - \mathbf{\tilde z}_{t_k}\|_2^2 \right)$
    \State $\tilde{\bm{\epsilon}}_{t_k} \leftarrow (\mathbf{\tilde x}_{t_k} - \alpha_{t_k} \hat{\mathbf{x}}_{0|t_k}) / \sigma_{t_k}$
    \State Sample $\bm \epsilon \sim \mathcal N(\bm 0, \mathbf I_N)$
    \State $\mathbf{\tilde x}_{t_{k-1}} \leftarrow \alpha_{t_{k-1}} \hat{\mathbf{x}}_{0|t_k} + \sigma_{t_{k-1}} (\sqrt{1-\zeta}\tilde{\bm{\epsilon}}_{t_k} + \sqrt{\zeta} \bm \epsilon)$
\EndFor
\State \Return $\mathbf{\tilde x}_{t_0}$
\end{algorithmic}
\end{algorithm}

To investigate this, we conducted experiments on the FFHQ validation sets using 100 images for the 1-bit CS task with an additive Gaussian noise level of $\sigma=0.5$. For these experiments, the penalty coefficient $\lambda$ was set to 0.05, and we varied $\zeta$ across the set $\{0, 0.2, 0.5, 0.8, 1.0\}$. The quantitative results are presented in Table~\ref{tab:zeta_ablation_full}.

The results demonstrate that for the 1-bit quantization task, a fully deterministic reverse process ($\zeta=0$) produces the highest quality reconstructions. Increasing the stochasticity ($\zeta > 0$) degrades the performance of Diff-OneBit.

\begin{table}[htbp]
\centering
\small
\begin{tabular}{l cccc}
\toprule
& \multicolumn{4}{c}{1-bit CS ($\sigma=0.5$)} \\
\cmidrule(lr){2-5}
\textbf{$\zeta$} & PSNR$\uparrow$ & SSIM$\uparrow$ & LPIPS$\downarrow$ & FID$\downarrow$ \\
\midrule
0 & $22.21\pm1.42$ & $0.649\pm0.07$ & $0.328\pm0.05$ & 85.40 \\
0.2 & $22.12\pm1.35$ & $0.630\pm0.07$ & $0.387\pm0.04$ & 107.67 \\
0.5 & $21.86\pm1.55$ & $0.633\pm0.07$ & $0.376\pm0.04$ & 111.47 \\
0.8 & $20.93\pm1.69$ & $0.606\pm0.08$ & $0.385\pm0.05$ & 108.68 \\
1.0 & $20.99\pm1.59$ & $0.595\pm0.07$ & $0.408\pm0.05$ & 117.22 \\
\bottomrule
\end{tabular}
\caption{{The performance of Diff-Onebit under different stochasticity level $\zeta$ for 1-bit CS on FFHQ images with $\sigma=0.5$.} We set $\lambda=0.05$ and NFE=20 for all the experiments.}
\label{tab:zeta_ablation_full}
\end{table}

\section{Ablation study on hyperparameters $\lambda$ and $\sigma$}

In this section, we present evaluation results for hyperparameters of our method. Regarding \(\lambda\), we conducted an ablation study with values of 0.01, 0.02, 0.05, and 0.1 on 1-bit CS using FFHQ images. Regarding \(\sigma\), we evaluated 1-bit CS using FFHQ images with values of 0.5, 1.0, 1.5, 2.0, and 2.5. The quantitative results are summarized in Table~\ref{tab:ablation_lambda} and Table~\ref{tab:ablation_sigma}, respectively.

\begin{table}[htbp]
\centering
\begin{tabular}{c c c c}
\toprule
\(\lambda\) & PSNR\(\uparrow\) & SSIM\(\uparrow\) & LPIPS\(\downarrow\) \\
\midrule
0.01 & 20.79 \(\pm\) 1.41 & 0.384 \(\pm\) 0.06 & 0.525 \(\pm\) 0.05 \\
0.02 & 22.05 \(\pm\) 1.88 & 0.642 \(\pm\) 0.07 & 0.341 \(\pm\) 0.05 \\
0.05 & 21.47 \(\pm\) 2.09 & 0.624 \(\pm\) 0.08 & 0.374 \(\pm\) 0.06 \\
0.1  & 20.65 \(\pm\) 2.32 & 0.571 \(\pm\) 0.09 & 0.448 \(\pm\) 0.06 \\
\bottomrule
\end{tabular}
\caption{Quantitative results of Diff-OneBit on FFHQ dataset under different \(\lambda\) values for 1-bit CS.}
\label{tab:ablation_lambda}
\end{table}

\begin{table}[htbp]
\centering
\begin{tabular}{c c c c}
\toprule
\(\sigma\) & PSNR\(\uparrow\) & SSIM\(\uparrow\) & LPIPS\(\downarrow\) \\
\midrule
0.5 & 22.05 \(\pm\) 1.88 & 0.642 \(\pm\) 0.07 & 0.341 \(\pm\) 0.05 \\
1.0 & 20.85 \(\pm\) 1.62 & 0.583 \(\pm\) 0.07 & 0.392 \(\pm\) 0.05 \\
1.5 & 20.25 \(\pm\) 1.01 & 0.532 \(\pm\) 0.06 & 0.441 \(\pm\) 0.05 \\
2.0 & 19.14 \(\pm\) 1.38 & 0.515 \(\pm\) 0.08 & 0.456 \(\pm\) 0.05 \\
2.5 & 16.49 \(\pm\) 1.37 & 0.497 \(\pm\) 0.09 & 0.484 \(\pm\) 0.08 \\
\bottomrule
\end{tabular}
\caption{Quantitative results of Diff-OneBit on FFHQ dataset under different \(\sigma\) values for 1-bit CS.}
\label{tab:ablation_sigma}
\end{table}

As shown in Table~\ref{tab:ablation_lambda}, \(\lambda=0.05\) achieves the best performance across all metrics, demonstrating an optimal balance between data consistency and regularization. Table~\ref{tab:ablation_sigma} indicates that smaller \(\sigma\) values lead to better reconstruction quality, with \(\sigma=1.0\) yielding the highest PSNR and SSIM scores while maintaining the lowest LPIPS.

\section{Computational efficiency}

In this section, we evaluate the computational efficiency of our proposed methods by comparing their inference times, a critical metric for practical applications. We compare Diff-OneBit with DPS, DAPS, DiffPIR, QCS-SGM and SIM-DMIS on the FFHQ validation sets in 1-bit CS with measurement ratio $M/N=1/16$, reporting the total time in seconds required to reconstruct 10 images. All experiments were conducted on a signal NVIDIA Tesla V100.

As presented in Table~\ref{tab:inference_time_ffhq}, SIM-DMIS is the most efficient method as expected. SIM-DMIS utilizes measurements in the beginning of an unconditional sampling process, thus avoiding costly iterative guidance. Our method, Diff-OneBit, ranks as the second, significantly outperforming other iterative guidance methods like DPS, DAPS, and DiffPIR. The efficiency of Diff-OneBit stems from its low NFE. However, each NFE involves 100 inner iteration steps to solve the optimization subproblem, which contributes to the overall runtime. This trade-off allows for strong guidance with minimal NFEs.  The results clearly highlight that our approach offers a compelling balance, providing a much more efficient solution than traditional guided methods.

\begin{table}[htbp]
\centering
\begin{tabular}{llc}
\toprule
\textbf{Method} & \textbf{NFE} & \textbf{Inference Time (s)} \\
\midrule
Diff-OneBit & 20   & {39.53}  \\
SIM-DMIS    & 150  & {5.64}   \\
DiffPIR     & 100  & 99.21  \\
DPS         & 1000 & 128    \\
DAPS        & 1000 & 425    \\
QCS-SGM  & 11555   & 1080  \\
\bottomrule
\end{tabular}
\caption{{Comparison of inference time (in seconds) for reconstructing 10 images on FFHQ.} We compare Diff-OneBit with DPS, DAPS, DIffPIR, SIM-DMIS and QCS-SGM for 1-bit CS on FFHQ images.}
\label{tab:inference_time_ffhq}
\end{table}

\section{Implementation details of our method and baselines}

In our experiments, we configure the hyperparameters for each method as follows.\\

\textbf{DDRM.} For 1-bit inpainting, we adopt the recommended values from~\cite{kawar2022denoising}, specifically \(\eta=0.85\) and \(\eta_b = 1.0\).\\

\textbf{DDNM.} Similarly, for 1-bit inpainting, we follow the recommended value from~\cite{wang2023zero}, setting \(\eta=0.85\).\\

\textbf{DPS.} We find that the learning rate \(\zeta\) recommended by~\cite{chung2023diffusion}, i.e., \(\zeta=0.3\), is too large for compressed sensing tasks. Therefore, we use a constant value \(\zeta= 0.1\) for both 1-bit CS and logistic regression.\\

\textbf{DAPS.} Following the recommended configuration from DAPS-1K in~\cite{zhang2025improving}, we set 5 ODE steps and 200 annealing steps. Additionally, we perform Langevin dynamics for 100 iterations per timestep with step size $\eta_t= \eta_0(\delta + t/T(1-\delta)),$
where \(\eta_0=1 \times 10^{-4}\), \(\delta=0.01\), and \(T=1000\).\\

\textbf{DiffPIR.} We employ the recommended values from~\cite{zhu2023denoising}, namely the DDIM hyperparameter \(\eta= 1.0\) and \(\lambda= 7.0\). Furthermore, we set the number of optimization steps to 50 and the learning rate to 0.1.\\

\textbf{QCS-SGM.} For 1-bit CS, we adopt the recommended values from~\cite{meng2023quantized}.\\

\textbf{SIM-DMIS.} We observe that the recommended values from~\cite{tang2025learning} are too small for \(3\times256\times256\) images. Consequently, we adjust the values from \(C_s=C_s'=55\) to \(C_s=C_s'=135\).\\

\textbf{Diff-OneBit.} We set \(\lambda=0.02\) for 1-bit CS and logistic regression tasks. For 1-bit inpainting, we increase this to \(\lambda=1\). Moreover, we configure the number of optimization steps to 100 and set the learning rate to 0.25.\\

\section{Out-of-distribution evaluation of diffusion priors}

To evaluate the generalization capability of diffusion priors across different data distributions, we conduct out-of-distribution (OOD) experiments. Specifically, we test our Diff-OneBit method on FFHQ validation images using a DM trained on the CelebA dataset, rather than on the FFHQ dataset. This setup assesses whether the learned diffusion prior can effectively regularize inverse problems even when the test distribution differs from the training distribution. In practice, we perform 1-bit CS reconstruction with Gaussian noise $\sigma=0.5$ on 100 FFHQ validation images, utilizing a DM that was pre-trained on the CelebA dataset. Table~\ref{tab:ood_evaluation} presents the quantitative results of OOD evaluation, demonstrating that the diffusion prior learned from CelebA transfers effectively to FFHQ images, despite the distribution shift between the two datasets.

\begin{table}[htbp]
\centering
\small
\begin{tabular}{l ccc}
\toprule
& \multicolumn{3}{c}{1-bit CS ($\sigma=0.5$)} \\
\cmidrule(lr){2-4}
\textbf{DM Training Data} & PSNR$\uparrow$ & SSIM$\uparrow$ & LPIPS$\downarrow$ \\
\midrule
FFHQ & $22.05\pm1.88$ & $0.64\pm0.07$ & $0.34\pm0.05$ \\
CelebA  & $21.76\pm1.90$ & $0.62\pm0.08$ & $0.35\pm0.06$ \\
\bottomrule
\end{tabular}
\caption{Out-of-distribution evaluation of diffusion priors. We compare the reconstruction performance on FFHQ validation images using DMs trained on FFHQ versus CelebA. The marginal performance gap demonstrates the generalization capability of diffusion priors across different image distributions.}
\label{tab:ood_evaluation}
\end{table}

\section{From MAP objective to alternating optimization with diffusion prior}

In the section, we provide a theoretical analysis of our approach, demonstrating the equivalence between the MAP objective in Eq.~\eqref{eq:map_objective} and the alternating optimization scheme, and deriving how the DM prior is incorporated through Tweedie's formula.\\
The HQS method reformulates the MAP objective from Eq.~\eqref{eq:map_objective} by introducing an auxiliary variable $\mathbf{z}$ and an equality constraint $\mathbf{x} = \mathbf{z}$:
\begin{equation}
\min_{\mathbf{x}, \mathbf{z}} \mathcal{L}(\mathbf{x}; \mathbf{y}) - \log p(\mathbf{z}) \quad \text{subject to} \quad \mathbf{x} = \mathbf{z}.
\end{equation}
Using the augmented Lagrangian method, the constrained problem is converted into an unconstrained form:
\begin{equation}
\min_{\mathbf{x}, \mathbf{z}} \mathcal{L}(\mathbf{x}; \mathbf{y}) - \log p(\mathbf{z}) + \frac{\mu}{2} \|\mathbf{x} - \mathbf{z}\|_2^2,
\end{equation}
where $\mu > 0$ is a penalty parameter that enforces the constraint $\mathbf{x} = \mathbf{z}$. The objective can be optimized by alternating between minimizing with respect to $\mathbf{x}$ and $\mathbf{z}$. At iteration $k$, given the current estimate $\hat{\mathbf{z}}^{(k)}$, we update $\mathbf{x}$ by solving:
\begin{equation}
\hat{\mathbf{x}}^{(k+1)} = \arg\min_{\mathbf{x}} \left( \mathcal{L}(\mathbf{x}; \mathbf{y}) + \frac{\mu}{2} \|\mathbf{x} - \hat{\mathbf{z}}^{(k)}\|_2^2 \right).
\end{equation}
This step enforces consistency with the observed measurements while maintaining proximity to the prior-regularized solution $\hat{\mathbf{z}}^{(k)}$. Next, for the updated estimate $\hat{\mathbf{x}}^{(k+1)}$, we update $\mathbf{z}$ by solving:
\begin{equation}
\hat{\mathbf{z}}^{(k+1)} = \arg\min_{\mathbf{z}} \left( -\log p(\mathbf{z}) + \frac{\mu}{2} \|\mathbf{z} - \hat{\mathbf{x}}^{(k+1)}\|_2^2 \right).
\end{equation}
The z-update can be solved using the DM prior. The optimization problem is equivalent to finding the MAP estimate under the posterior:
\begin{equation}
p(\mathbf{z} | \hat{\mathbf{x}}^{(k+1)}) \propto p(\mathbf{z}) \exp\left(-\frac{\mu}{2} \|\mathbf{z} - \hat{\mathbf{x}}^{(k+1)}\|_2^2\right).
\end{equation}
To establish the connection with the DM framework, we leverage the forward diffusion process. In the DM framework, a noisy observation $\mathbf{x}_t$ at diffusion step $t$ relates to the clean image $\mathbf{x}_0$ through:
\begin{equation}
\mathbf{x}_t = \alpha_t \mathbf{x}_0 + \sigma_t \bm{\epsilon}, \quad \bm{\epsilon} \sim \mathcal{N}(\mathbf{0}, \mathbf{I}),
\end{equation}
which implies $\mathbf{x}_t | \mathbf{x}_0 \sim \mathcal{N}(\alpha_t \mathbf{x}_0, \sigma_t^2 \mathbf{I})$. The corresponding likelihood is:
\begin{equation}
p(\mathbf{x}_t | \mathbf{x}_0) \propto \exp\left(-\frac{1}{2\sigma_t^2} \|\mathbf{x}_t - \alpha_t \mathbf{x}_0\|_2^2\right).
\end{equation}
To establish an exact correspondence with the HQS posterior, we introduce a rescaled variable: $\tilde{\mathbf{x}}_t = {\mathbf{x}_t}/{\alpha_t}$. Substituting into the likelihood, we obtain:
\begin{equation}
\|\mathbf{x}_t - \alpha_t \mathbf{x}_0\|_2^2 = \|\alpha_t \tilde{\mathbf{x}}_t - \alpha_t \mathbf{x}_0\|_2^2 = \alpha_t^2 \|\tilde{\mathbf{x}}_t - \mathbf{x}_0\|_2^2.
\end{equation}
Thus, the likelihood becomes:
\begin{equation}
p(\mathbf{x}_t | \mathbf{x}_0) \propto \exp\left(-\frac{\alpha_t^2}{2\sigma_t^2} \|\tilde{\mathbf{x}}_t - \mathbf{x}_0\|_2^2\right).
\end{equation}
Now, by identifying $\tilde{\mathbf{x}}_t = \hat{\mathbf{x}}^{(k+1)}$ and $\mathbf{x}_0 = \mathbf{z}$, the DM posterior can be written as:
\begin{equation}
p(\mathbf{z} | \tilde{\mathbf{x}}_t) \propto p(\mathbf{z}) \cdot p(\tilde{\mathbf{x}}_t | \mathbf{z}) \propto p(\mathbf{z}) \exp\left(-\frac{\alpha_t^2}{2\sigma_t^2} \|\tilde{\mathbf{x}}_t - \mathbf{z}\|_2^2\right).
\end{equation}
Comparing this with the HQS posterior:
\begin{equation}
p(\mathbf{z} | \hat{\mathbf{x}}^{(k+1)}) \propto p(\mathbf{z}) \exp\left(-\frac{\mu}{2} \|\hat{\mathbf{x}}^{(k+1)} - \mathbf{z}\|_2^2\right),
\end{equation}
we obtain an exact correspondence by setting: $\mu = {\alpha_t^2}/{\sigma_t^2}$. In practice, we set $\mu = \lambda {\alpha_t^2}/{\sigma_t^2}$, where $\lambda$ is a hyperparameter controlling the strength of the prior regularization. With this reparameterization, the z-update becomes equivalent to estimating the clean image $\mathbf{x}_0$ from the rescaled observation $\tilde{\mathbf{x}}_t = {\mathbf{x}_t}/{\alpha_t}$ in the DM framework. Therefore, we apply Tweedie's formula, which provides the posterior mean estimate:
\begin{equation}
\mathbb{E}[\mathbf{x}_0 | \mathbf{x}_t] = \frac{\mathbf{x}_t - \sigma_t \bm{\epsilon}_{\btheta}(\mathbf{x}_t, t)}{\alpha_t},
\end{equation}
where $\bm{\epsilon}_{\btheta}(\mathbf{x}_t, t)$ is the noise prediction from the pre-trained DM. Since $\tilde{\mathbf{x}}_t = \hat{\mathbf{x}}^{(k+1)}$, the corresponding unscaled observation is $\mathbf{x}_t = \alpha_t \hat{\mathbf{x}}^{(k+1)}$. Substituting this into Tweedie's formula yields the z-update:
\begin{equation}
\label{eq:tweeide_z_update}
\hat{\mathbf{z}}^{(k+1)} = \mathbb{E}[\mathbf{x}_0 | \mathbf{x}_t = \alpha_t \hat{\mathbf{x}}^{(k+1)}] = \frac{\alpha_t \hat{\mathbf{x}}^{(k+1)} - \sigma_t \bm{\epsilon}_{\btheta}(\alpha_t \hat{\mathbf{x}}^{(k+1)}, t)}{\alpha_t}.
\end{equation}
Eq.~\eqref{eq:tweeide_z_update} establishes the relationship between the clean image estimate $\hat{\mathbf{z}}^{(k+1)}$ and the unscaled observation $\mathbf{x}_t = \alpha_t \hat{\mathbf{x}}^{(k+1)}$. The predicted clean image $\hat{\mathbf{z}}^{(k+1)}$ inherently incorporates the learned image prior from the DM through the noise prediction network $\bm{\epsilon}_{\btheta}$, and serves as the prior-regularized estimate for the next x-update. As the penalty parameter $\mu$ increases through the reverse diffusion process, the constraint $\mathbf{x} = \mathbf{z}$ is increasingly enforced, and the alternating scheme converges to a solution that balances data fidelity with the diffusion prior, thereby solving the original MAP estimation problem in Eq.~\eqref{eq:map_objective}.

\section{Norm estimation} 
\begin{figure}[htbp]
    \centering
    \includegraphics[width=1\linewidth]{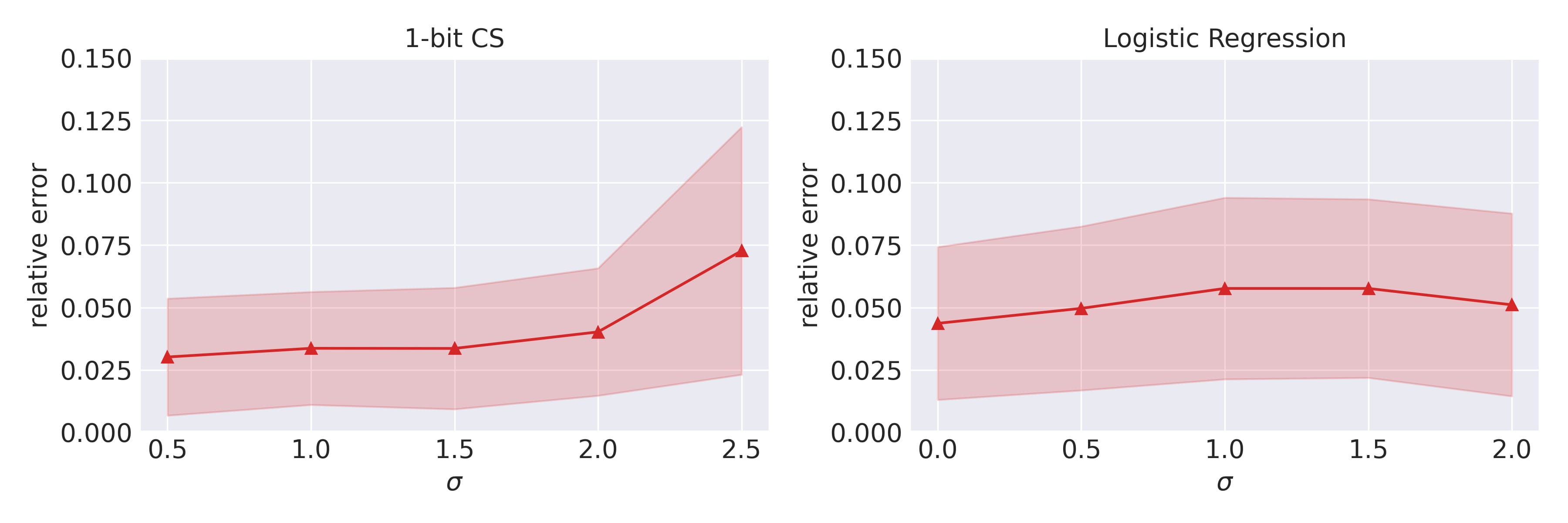}
    \caption{Norm estimation. We compare Diff-OneBit for 1-bit CS and logistic regression on FFHQ images with varying noise level $\sigma$.}
    \label{fig:dual_axis_metrics_plot}
\end{figure}

Under 1-bit quantization, it is typically difficult to estimate the norm of the underlying signal~\cite{knudson2016one}. To demonstrate the capability of Diff-OneBit for norm estimation, we evaluated Diff-OneBit across varying levels of Gaussian noise in 1-bit CS and logistic regression on the FFHQ dataset. We tested pre-quantization noise levels at $\sigma \in \{0.5, 1.0, 1.5, 2.0, 2.5\}$ for 1-bit CS and $\sigma \in \{0, 0.5, 1.0, 1.5, 2.0\}$ for logistic regression. In logistic regression, Diff-OneBit is designed for the noiseless scenario with $\sigma=0$, and we simply reused this method for norm estimation. As shown in Figure~\ref{fig:dual_axis_metrics_plot}, our method generally keeps the relative error between the $\ell_2$-norms of the recovered vector $\tilde{\mathbf{x}}$ and the underlying signal $\mathbf{x}^*$, defined as ${| \|\tilde{\mathbf{x}}\|_2 - \|\mathbf{x}^*\|_2 |} /{\|\mathbf{x}^*\|_2}$, within a small range.

\begin{figure}[htbp]
    \centering
    \includegraphics[width=1\linewidth]{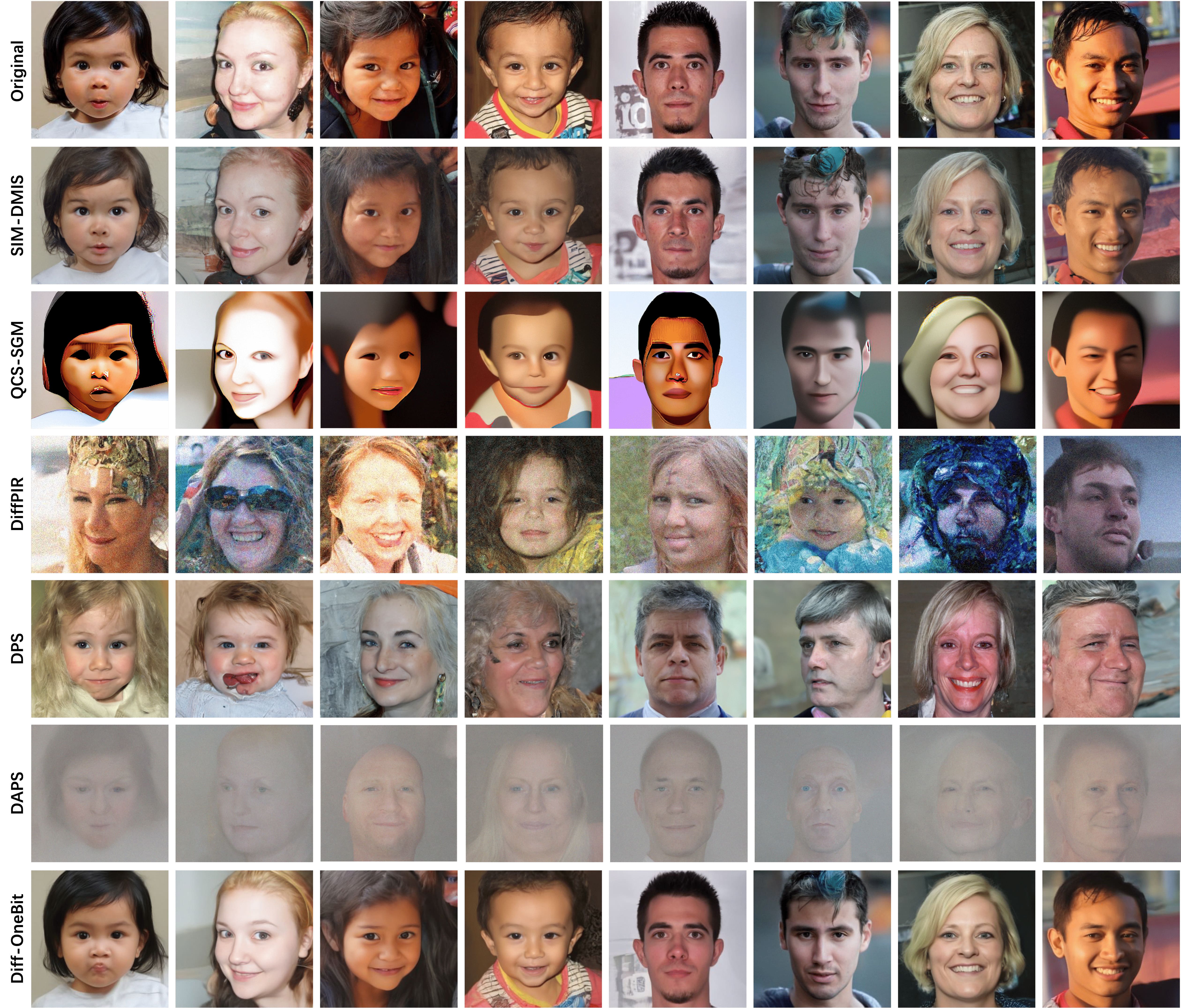}
    \caption{Additional visual results for 1-bit CS on the FFHQ images with $\sigma=0.5$.}
    \label{fig:appendix_ffhq_1bitcs_methods}
\end{figure}

\begin{figure}[htbp]
    \centering
    \includegraphics[width=1\linewidth]{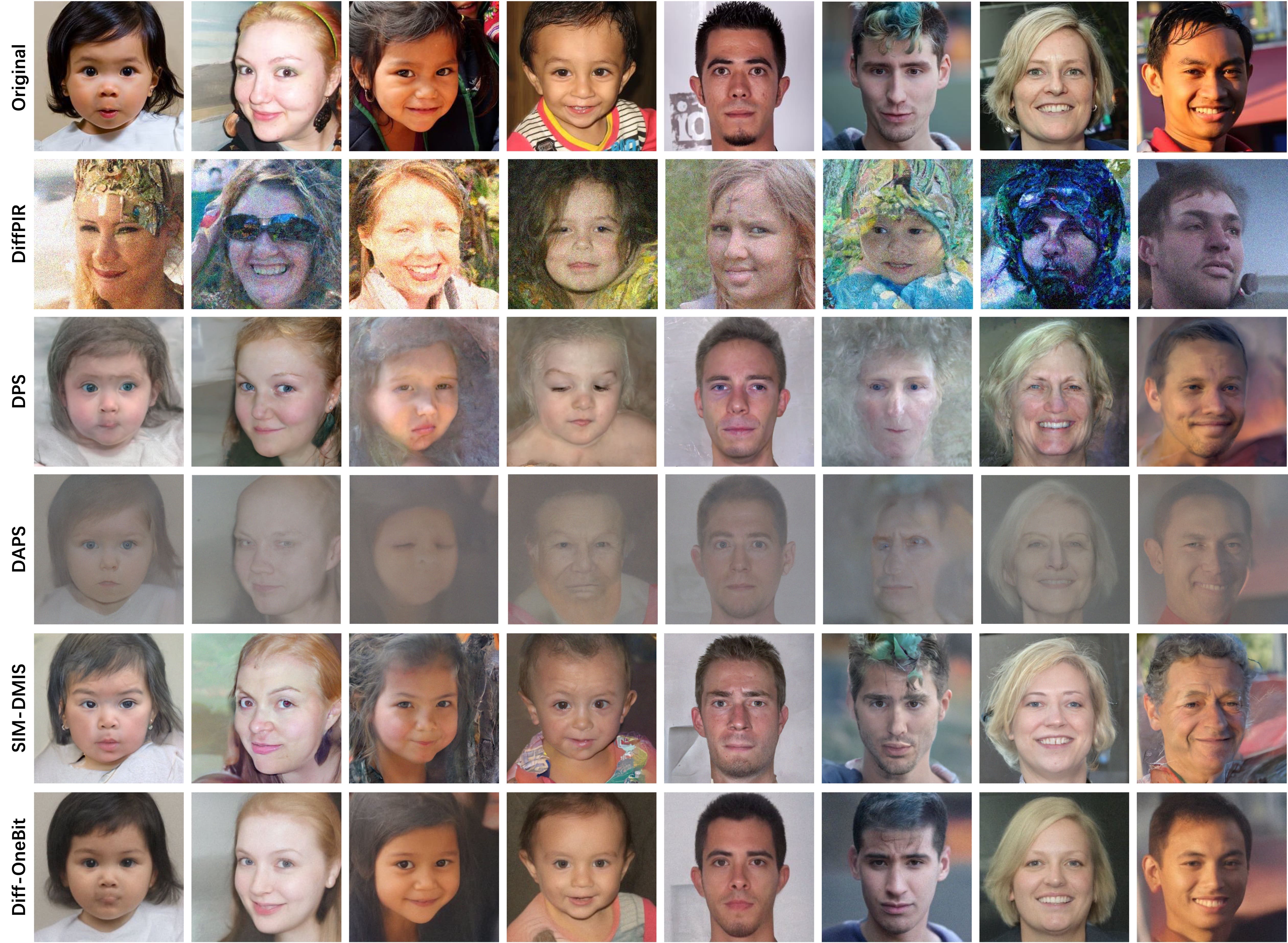}
    \caption{Additional visual results for logistic regression on FFHQ images with $\sigma=0$.}
    \label{fig:appendix_ffhq_lr_methods}
\end{figure}

\bibliographystyle{IEEEtran}
\bibliography{aaai2026}

@article{chan2016plug,
  title={Plug-and-play {ADMM} for image restoration: Fixed-point convergence and applications},
  author={Chan, Stanley H and Wang, Xiran and Elgendy, Omar A},
  journal={IEEE Transactions on Computational Imaging},
  volume={3},
  number={1},
  pages={84--98},
  year={2016}
}

@article{kamilov2017plug,
  title={A plug-and-play priors approach for solving nonlinear imaging inverse problems},
  author={Kamilov, Ulugbek S and Mansour, Hassan and Wohlberg, Brendt},
  journal={IEEE Signal Processing Letters},
  volume={24},
  number={12},
  pages={1872--1876},
  year={2017}
}

@article{sreehari2016plug,
  title={Plug-and-play priors for bright field electron tomography and sparse interpolation},
  author={Sreehari, Suhas and Venkatakrishnan, S Venkat and Wohlberg, Brendt and Buzzard, Gregery T and Drummy, Lawrence F and Simmons, Jeffrey P and Bouman, Charles A},
  journal={IEEE Transactions on Computational Imaging},
  volume={2},
  number={4},
  pages={408--423},
  year={2016}
}

@inproceedings{venkatakrishnan2013plug,
  title={Plug-and-play priors for model based reconstruction},
  author={Venkatakrishnan, Singanallur V and Bouman, Charles A and Wohlberg, Brendt},
  booktitle={GlobalSIP},
  year={2013}
}

@inproceedings{wei2019statistical,
  title={On the statistical rate of nonlinear recovery in generative models with heavy-tailed data},
  author={Wei, Xiaohan and Yang, Zhuoran and Wang, Zhaoran},
  booktitle={ICML},
  year={2019}
}

@inproceedings{zheng2025integrating,
  title={Integrating Intermediate Layer Optimization and Projected Gradient Descent for Solving Inverse Problems with Diffusion Models},
  author={Zheng, Yang and Li, Wen and Liu, Zhaoqiang},
  booktitle={ICML},
year={2025}
}

@article{chen2023unified,
  title={A unified framework for uniform signal recovery in nonlinear generative compressed sensing},
  author={Chen, Junren and Scarlett, Jonathan and Ng, Michael and Liu, Zhaoqiang},
  journal={NeurIPS},
  year={2023}
}

@inproceedings{liu2022non,
  title={Non-iterative recovery from nonlinear observations using generative models},
  author={Liu, Jiulong and Liu, Zhaoqiang},
  booktitle={CVPR},
  year={2022}
}

@article{liu2020generalized,
  title={The generalized {L}asso with nonlinear observations and generative priors},
  author={Liu, Zhaoqiang and Scarlett, Jonathan},
  journal={NeurIPS},
  year={2020}
}

@article{plan2013one,
  title={One-bit compressed sensing by linear programming},
  author={Plan, Yaniv and Vershynin, Roman},
  journal={Communications on Pure and Applied Mathematics},
  volume={66},
  number={8},
  pages={1275--1297},
  year={2013}
}

@article{plan2017high,
  title={High-dimensional estimation with geometric constraints},
  author={Plan, Yaniv and Vershynin, Roman and Yudovina, Elena},
  journal={Information and Inference: A Journal of the IMA},
  volume={6},
  number={1},
  pages={1--40},
  year={2017}
}

@article{knudson2016one,
  title={One-bit compressive sensing with norm estimation},
  author={Knudson, Karin and Saab, Rayan and Ward, Rachel},
  journal={IEEE Transactions on Information Theory},
  volume={62},
  number={5},
  pages={2748--2758},
  year={2016}
}

@article{davenport20141,
  title={1-bit matrix completion},
  author={Davenport, Mark A and Plan, Yaniv and Van Den Berg, Ewout and Wootters, Mary},
  journal={Information and Inference: A Journal of the IMA},
  volume={3},
  number={3},
  pages={189--223},
  year={2014}
}

@article{plan2012robust,
  title={Robust 1-bit compressed sensing and sparse logistic regression: A convex programming approach},
  author={Plan, Yaniv and Vershynin, Roman},
  journal={IEEE Transactions on Information Theory},
  volume={59},
  number={1},
  pages={482--494},
  year={2012}
}

@article{yan2012robust,
  title={Robust 1-bit compressive sensing using adaptive outlier pursuit},
  author={Yan, Ming and Yang, Yi and Osher, Stanley},
  journal={IEEE Transactions on Signal Processing},
  volume={60},
  number={7},
  pages={3868--3875},
  year={2012}
}

@article{jacques2013robust,
  title={Robust 1-bit compressive sensing via binary stable embeddings of sparse vectors},
  author={Jacques, Laurent and Laska, Jason N and Boufounos, Petros T and Baraniuk, Richard G},
  journal={IEEE Transactions on Information Theory},
  volume={59},
  number={4},
  pages={2082--2102},
  year={2013}
}

@inproceedings{boufounos20081,
  title={1-bit compressive sensing},
  author={Boufounos, Petros T and Baraniuk, Richard G},
  booktitle={CISS},
  year={2008}
}

@inproceedings{boufounos2010reconstruction,
  title={Reconstruction of sparse signals from distorted randomized measurements},
  author={Boufounos, Petros T},
  booktitle={ICASSP},
  year={2010}
}

@inproceedings{ho2020denoising,
    author = "Ho, Jonathan and Jain, Ajay and Abbeel, Pieter",
    title = "Denoising diffusion probabilistic models",
    booktitle = "NeurIPS",
    year = 2020
}

@inproceedings{karras2022elucidating,
    author = "Karras, Tero and Aittala, Miika and Aila, Timo and Laine, Samuli",
    title = "Elucidating the design space of diffusion-based generative models",
    booktitle = "NeurIPS",
    year = 2022
}

@inproceedings{song2019generative,
  title={Generative modeling by estimating gradients of the data distribution},
  author={Song, Yang and Ermon, Stefano},
  booktitle={NeurIPS},
  year={2019}
}

@inproceedings{song2020improved,
  title={Improved techniques for training score-based generative models},
  author={Song, Yang and Ermon, Stefano},
  booktitle={NeurIPS},
  year={2020}
}

@inproceedings{song2021score,
  title={Score-Based Generative Modeling through Stochastic Differential Equations},
  author={Song, Yang and Sohl-Dickstein, Jascha and Kingma, Diederik P and Kumar, Abhishek and Ermon, Stefano and Poole, Ben},
  booktitle={ICLR},
  year={2021}
}

@inproceedings{sohl2015deep,
  title={Deep unsupervised learning using nonequilibrium thermodynamics},
  author={Sohl-Dickstein, Jascha and Weiss, Eric and Maheswaranathan, Niru and Ganguli, Surya},
  booktitle={ICML},
  year={2015}
}

@article{liu2020information,
  title={Information-theoretic lower bounds for compressive sensing with generative models},
  author={Liu, Zhaoqiang and Scarlett, Jonathan},
  journal={IEEE Journal on Selected Areas in Information Theory},
  volume={1},
  number={1},
  pages={292--303},
  year={2020}
}

@inproceedings{liu2020sample,
  title={Sample complexity bounds for 1-bit compressive sensing and binary stable embeddings with generative priors},
  author={Liu, Zhaoqiang and Gomes, Selwyn and Tiwari, Avtansh and Scarlett, Jonathan},
  booktitle={ICML},
  year={2020}
}

@inproceedings{qiu2020robust,
  title={Robust one-bit recovery via {ReLU} generative networks: Near-optimal statistical rate and global landscape analysis},
  author={Qiu, Shuang and Wei, Xiaohan and Yang, Zhuoran},
  booktitle={ICML},
  year={2020}
}

@article{daras2024survey,
  title={A survey on diffusion models for inverse problems},
  author={Daras, Giannis and Chung, Hyungjin and Lai, Chieh-Hsin and Mitsufuji, Yuki and Ye, Jong Chul and Milanfar, Peyman and Dimakis, Alexandros G and Delbracio, Mauricio},
  journal={https://arxiv.org/2410.00083},
  year={2024}
}

@inproceedings{song2021denoising,
  title={Denoising Diffusion Implicit Models},
  author={Song, Jiaming and Meng, Chenlin and Ermon, Stefano},
  booktitle={ICLR},
  year={2021}
}

@inproceedings{chung2023diffusion,
  title={Diffusion Posterior Sampling for General Noisy Inverse Problems},
  author={Chung, Hyungjin and Kim, Jeongsol and McCann, Michael T and Klasky, Marc L and Ye, Jong Chul},
  booktitle={ICLR},
  year={2023}
}

@inproceedings{karras2019style,
  title={A style-based generator architecture for generative adversarial networks},
  author={Karras, Tero and Laine, Samuli and Aila, Timo},
  booktitle={CVPR},
  year={2019}
}

@inproceedings{karras2018progressive,
  title={Progressive Growing of {GAN}s for Improved Quality, Stability, and Variation},
  author={Karras, Tero and Aila, Timo and Laine, Samuli and Lehtinen, Jaakko},
  booktitle={ICLR},
  year={2018}
}

@inproceedings{dhariwal2021diffusion,
  title={Diffusion models beat {GAN}s on image synthesis},
  author={Dhariwal, Prafulla and Nichol, Alexander},
  booktitle={NeurIPS},
  year={2021}
}

@inproceedings{deng2009imagenet,
  title={Image{N}et: A large-scale hierarchical image database},
  author={Deng, Jia and Dong, Wei and Socher, Richard and Li, Li-Jia and Li, Kai and Fei-Fei, Li},
  booktitle={CVPR},
  year={2009}
}

@inproceedings{meng2023quantized,
  title={Quantized Compressed Sensing with Score-Based Generative Models},
  author={Meng, Xiangming and Kabashima, Yoshiyuki},
  booktitle={ICLR},
  year={2023},
}

@inproceedings{tang2025learning,
  title={Learning Single Index Models with Diffusion Priors},
  author={Tang, Anqi and Chen, Youming and Xue, Shuchen and Liu, Zhaoqiang},
  booktitle={ICML},
  year={2025}
}

@inproceedings{zhu2023denoising,
  title={Denoising diffusion models for plug-and-play image restoration},
  author={Zhu, Yuanzhi and Zhang, Kai and Liang, Jingyun and Cao, Jiezhang and Wen, Bihan and Timofte, Radu and Van Gool, Luc},
  booktitle={CVPRW},
  year={2023}
}

@article{kingma2014adam,
  title={Adam: A method for stochastic optimization},
  author={Kingma, Diederik P and Ba, Jimmy},
  journal={https://arxiv.org/1412.6980},
  year={2014}
}

@inproceedings{zhang2018unreasonable,
  title={The unreasonable effectiveness of deep features as a perceptual metric},
  author={Zhang, Richard and Isola, Phillip and Efros, Alexei A and Shechtman, Eli and Wang, Oliver},
  booktitle={CVPR},
  year={2018}
}

@article{heusel2017gans,
  title={G{AN}s trained by a two time-scale update rule converge to a local {N}ash equilibrium},
  author={Heusel, Martin and Ramsauer, Hubert and Unterthiner, Thomas and Nessler, Bernhard and Hochreiter, Sepp},
  journal={NeurIPS},
  year={2017}
}

@article{candes2006robust,
  title={Robust uncertainty principles: Exact signal reconstruction from highly incomplete frequency information},
  author={Cand{\`e}s, Emmanuel J and Romberg, Justin and Tao, Terence},
  journal={IEEE Transactions on Information Theory},
  volume={52},
  number={2},
  pages={489--509},
  year={2006}
}

@article{donoho2006compressed,
  title={Compressed sensing},
  author={Donoho, David L},
  journal={IEEE Transactions on Information Theory},
  volume={52},
  number={4},
  pages={1289--1306},
  year={2006}
}

@incollection{foucart2013invitation,
  title={An invitation to compressive sensing},
  author={Foucart, Simon and Rauhut, Holger},
  booktitle={A mathematical introduction to compressive sensing},
  pages={1--39},
  year={2013}
}

@article{chen2013low,
  title={Low-rank matrix recovery from errors and erasures},
  author={Chen, Yudong and Jalali, Ali and Sanghavi, Sujay and Caramanis, Constantine},
  journal={IEEE Transactions on Information Theory},
  volume={59},
  number={7},
  pages={4324--4337},
  year={2013}
}

@article{li2013compressed,
  title={Compressed sensing and matrix completion with constant proportion of corruptions},
  author={Li, Xiaodong},
  journal={Constructive Approximation},
  volume={37},
  number={1},
  pages={73--99},
  year={2013}
}

@article{nguyen2012robust,
  title={Robust {L}asso with missing and grossly corrupted observations},
  author={Nguyen, Nam H and Tran, Trac D},
  journal={IEEE Transactions on Information Theory},
  volume={59},
  number={4},
  pages={2036--2058},
  year={2012}
}

@article{nguyen2013exact,
  title={Exact Recoverability From Dense Corrupted Observations via {$\ell_{1}$}-{Minimization}},
  author={Nguyen, Nam H. and Tran, Trac D.},
  journal={IEEE Transactions on Information Theory},
  volume={59},
  number={4},
  pages={2017--2035},
  year={2013}
}

@article{xu2012outlier,
  title={Outlier-robust {PCA}: The high-dimensional case},
  author={Xu, Huan and Caramanis, Constantine and Mannor, Shie},
  journal={IEEE Transactions on Information Theory},
  volume={59},
  number={1},
  pages={546--572},
  year={2012}
}

@article{kawar2022denoising,
  title={Denoising diffusion restoration models},
  author={Kawar, Bahjat and Elad, Michael and Ermon, Stefano and Song, Jiaming},
  journal={NeurIPS},
  year={2022}
}

@inproceedings{bora2017compressed,
  title={Compressed sensing using generative models},
  author={Bora, Ashish and Jalal, Ajil and Price, Eric and Dimakis, Alexandros G},
  booktitle={ICML},
  year={2017}
}

@inproceedings{wang2023zero,
  title={Zero-Shot Image Restoration Using Denoising Diffusion Null-Space Model},
  author={Wang, Yinhuai and Yu, Jiwen and Zhang, Jian},
  booktitle={ICLR},
  year={2023},
}

@inproceedings{song2024solving,
    title={Solving Inverse Problems with Latent Diffusion Models via Hard Data Consistency},
    author={Bowen Song and Soo Min Kwon and Zecheng Zhang and Xinyu Hu and Qing Qu and Liyue Shen},
    booktitle={ICLR},
    year={2024},
}

@inproceedings{chung2022improving,
  title={Improving diffusion models for inverse problems using manifold constraints},
  author={Chung, Hyungjin and Sim, Byeongsu and Ryu, Dohoon and Ye, Jong Chul},
  booktitle={NeurIPS},
  year={2022}
}

@inproceedings{jalal2021robust,
  title={Robust compressed sensing {MRI} with deep generative priors},
  author={Jalal, Ajil and Arvinte, Marius and Daras, Giannis and Price, Eric and Dimakis, Alexandros G and Tamir, Jon},
  booktitle={NeurIPS},
  year={2021}
}

@inproceedings{choi2021ilvr,
  title={I{LVR}: Conditioning Method for Denoising Diffusion Probabilistic Models},
  author={Choi, Jooyoung and Kim, Sungwon and Jeong, Yonghyun and Gwon, Youngjune and Yoon, Sungroh},
  booktitle={ICCV},
  year={2021}
}

@inproceedings{kadkhodaie2021stochastic,
  title={Stochastic solutions for linear inverse problems using the prior implicit in a denoiser},
  author={Kadkhodaie, Zahra and Simoncelli, Eero},
  booktitle={NeurIPS},
  year={2021}
}

@article{geman1995nonlinear,
  title={Nonlinear image recovery with half-quadratic regularization},
  author={Geman, Donald and Yang, Chengda},
  journal={IEEE Transactions on Image Processing},
  volume={4},
  number={7},
  pages={932--946},
  year={1995}
}

@inproceedings{mardani2024variational,
  title={A Variational Perspective on Solving Inverse Problems with Diffusion Models},
  author={Mardani, Morteza and Song, Jiaming and Kautz, Jan and Vahdat, Arash},
  booktitle={ICLR},
  year={2024}
}

@inproceedings{song2023pseudoinverse,
  title={Pseudoinverse-guided diffusion models for inverse problems},
  author={Song, Jiaming and Vahdat, Arash and Mardani, Morteza and Kautz, Jan},
  booktitle={ICLR},
  year={2023}
}

@inproceedings{zhang2025improving,
  title={Improving diffusion inverse problem solving with decoupled noise annealing},
  author={Zhang, Bingliang and Chu, Wenda and Berner, Julius and Meng, Chenlin and Anandkumar, Anima and Song, Yang},
  booktitle={CVPR},
  year={2025}
}

@article{van2018compressed,
  title={Compressed sensing with deep image prior and learned regularization},
  author={Van Veen, Dave and Jalal, Ajil and Soltanolkotabi, Mahdi and Price, Eric and Vishwanath, Sriram and Dimakis, Alexandros G},
  journal={https://arxiv.org/1806.06438},
  year={2018}
}

@inproceedings{heckel2019deep,
  title={Deep Decoder: Concise Image Representations from Untrained Non-convolutional Networks},
  author={Heckel, Reinhard and Hand, Paul},
  booktitle={ICLR},
  year={2019}
}

@inproceedings{asim2020invertible,
  title={Invertible generative models for inverse problems: {M}itigating representation error and dataset bias},
  author={Asim, Muhammad and Daniels, Max and Leong, Oscar and Ahmed, Ali and Hand, Paul},
  booktitle={ICML},
  year={2020}
}

@article{ongie2020deep,
  title={Deep learning techniques for inverse problems in imaging},
  author={Ongie, Gregory and Jalal, Ajil and Metzler, Christopher A and others},
  journal={IEEE Journal on Selected Areas in Information Theory},
  volume={1},
  number={1},
  pages={39--56},
  year={2020}
}

@article{scarlett2023theoretical,
  title={Theoretical perspectives on deep learning methods in inverse problems},
  author={Scarlett, Jonathan and Heckel, Reinhard and Rodrigues, Miguel RD and Hand, Paul and Eldar, Yonina C},
  journal={IEEE Journal on Selected Areas in Information Theory},
  volume={3},
  number={3},
  pages={433--453},
  year={2023}
}

@inproceedings{jalal2021instance,
  title={Instance-Optimal Compressed Sensing via Posterior Sampling},
  author={Jalal, A and Karmalkar, S and Dimakis, A and Price, E},
  booktitle={ICML},
  year={2021}
}

@article{nguyen2021provable,
  title={Provable Compressed Sensing with Generative Priors via {L}angevin Dynamics},
  author={Nguyen, Thanh V and Jagatap, Gauri and Hegde, Chinmay},
  journal={https://arxiv.org/2102.12643},
  year={2021}
}

@inproceedings{liu2021towards,
  title={Towards Sample-Optimal Compressive Phase Retrieval with Sparse and Generative Priors},
  author={Liu, Zhaoqiang and Ghosh, Subhroshekhar and Scarlett, Jonathan},
  booktitle={NeurIPS},
    year={2021}
}

@inproceedings{liu2022generative,
title={Generative Principal Component Analysis},
author={Zhaoqiang Liu and Jiulong Liu and Subhroshekhar Ghosh and Jun Han and Jonathan Scarlett},
booktitle={ICLR},
year={2022}
}

@article{genzel2022solving,
  title={Solving inverse problems with deep neural networks--robustness included?},
  author={Genzel, Martin and Macdonald, Jan and M{\"a}rz, Maximilian},
  journal={IEEE Transactions on Pattern Analysis and Machine Intelligence},
  volume={45},
  number={1},
  pages={1119--1134},
  year={2022}
}

@inproceedings{liu2024generalized,
  title={Generalized Eigenvalue Problems with Generative Priors},
  author={Liu, Zhaoqiang and Li, Wen and Chen, Junren},
  booktitle={NeurIPS},
  year={2024}
}

@inproceedings{liu2022projected,
  title={Projected gradient descent algorithms for solving nonlinear inverse problems with generative priors},
  author={Liu, Zhaoqiang and Han, Jun},
  booktitle={IJCAI},
  year={2022}
}

@inproceedings{liu2022misspecified,
  title={Misspecified phase retrieval with generative priors},
  author={Liu, Zhaoqiang and Wang, Xinshao and Liu, Jiulong},
  booktitle={NeurIPS},
  year={2022}
}

@inproceedings{hand2018phase,
  title={Phase retrieval under a generative prior},
  author={Hand, Paul and Leong, Oscar and Voroninski, Vlad},
  booktitle={NeurIPS},
  year={2018}
}

@article{chen2025solving,
  title={Solving Quadratic Systems With Full-Rank Matrices Using Sparse or Generative Priors},
  author={Chen, Junren and Ng, Michael K and Liu, Zhaoqiang},
  journal={IEEE Transactions on Signal Processing},
  year={2025}
}

\end{document}